\PassOptionsToPackage{table}{xcolor}
\documentclass{eitnlp}

\usepackage{bm}
\usepackage{wrapfig}
\usepackage{algorithm}
\usepackage{algpseudocode}
\usepackage{multirow}
\usepackage{adjustbox}
\usepackage{makecell}
\usepackage{tabularx}
\usepackage{placeins}
\usepackage{printlen}
\usepackage{pifont}
\usepackage{tikz}
\graphicspath{{Figure/}}

\newsavebox{\tblbox}
\newcommand{\cnum}[1]{\ding{\the\numexpr171+#1\relax}}

\definecolor{sinkRaw}{HTML}{3C4048}
\definecolor{sinkCtl}{HTML}{B3B3B3}
\definecolor{sinkPale}{HTML}{C9C9C9}
\definecolor{sinkTop}{HTML}{D55E00}
\definecolor{gtBox}{HTML}{2ECC8F}
\definecolor{mapRaw}{HTML}{CC6677}
\definecolor{mapSm}{HTML}{4477AA}
\definecolor{mapRes}{HTML}{8C8C8C}
\definecolor{routeImg}{HTML}{E69F00}
\definecolor{routeTxt}{HTML}{0072B2}
\definecolor{routeBoth}{HTML}{009E73}
\definecolor{routeRand}{HTML}{999999}
\providecommand{\sw}[1]{{\color{#1}\rule[-0.15ex]{1.3ex}{1.3ex}}}
\providecommand{\legseg}[2]{{\color{#1}$\vcenter{\hrule height0.7pt width#2}$}}
\providecommand{\legmk}[2]{{\color{#1}$\vcenter{\hbox{$#2$}}$}}
\providecommand{\capleg}[3]{\mbox{%
  \legseg{#1}{0.5em}\kern-0.06em\legmk{#1}{#2}\kern-0.06em
  \legseg{#1}{0.5em}\,#3}}
\providecommand{\caplegd}[3]{\mbox{%
  \legseg{#1}{0.2em}\kern0.13em\legseg{#1}{0.2em}\kern-0.06em
  \legmk{#1}{#2}\kern-0.06em
  \legseg{#1}{0.2em}\kern0.13em\legseg{#1}{0.2em}\,#3}}
\definecolor{tierbg}{HTML}{F0F0F0}
\definecolor{oursbg}{HTML}{EAF1FB}
\definecolor{catGen}{HTML}{2A6F97}
\definecolor{catRea}{HTML}{2E8B57}
\definecolor{catOcr}{HTML}{C0662B}
\definecolor{catHal}{HTML}{8E5A9B}
\definecolor{refA}{HTML}{1F7A7A}
\definecolor{refB}{HTML}{3F5BA9}
\definecolor{refC}{HTML}{A8475A}
\newcommand{\gd}[2]{\cellcolor{#1!13}\textcolor{#1}{#2}}
\newcommand{\cat}[3]{\multicolumn{#1}{>{\columncolor{#2!13}}c}{\textcolor{#2}{\textbf{#3}}}}
\definecolor{keyq}{HTML}{C0392B}
\newcommand{\keyq}[1]{{\color{keyq}\textbf{\textit{#1}}}}
\definecolor{apxc}{HTML}{1F5A7A}
\newcommand{\apx}[1]{{\color{apxc}\textbf{#1}}}
\newcommand{\ub}[1]{\textcolor{black!55}{#1}}
\definecolor{mkFine}{HTML}{FF8C00}
\definecolor{mkFineEdge}{HTML}{5C2A00}
\definecolor{mkCoarse}{HTML}{FFD633}
\definecolor{mkGT}{HTML}{2ECC40}
\definecolor{mkPred}{HTML}{FF4D4D}
\definecolor{mkTileA}{HTML}{6FB3FF}
\definecolor{mkTileB}{HTML}{FF7F7F}
\DeclareRobustCommand{\mkgt}{\tikz[baseline=-0.6ex]\draw[mkGT,line width=0.9pt]
  (-0.75ex,-0.55ex) rectangle (0.75ex,0.55ex);}
\DeclareRobustCommand{\mkpred}{\tikz[baseline=-0.6ex]\draw[mkPred,line width=0.9pt]
  (-0.75ex,-0.55ex) rectangle (0.75ex,0.55ex);}
\DeclareRobustCommand{\mkfine}{\tikz[baseline=-0.6ex]\draw[fill=mkFine,
  draw=mkFineEdge,line width=0.3pt] (0,0) circle (0.5ex);}
\DeclareRobustCommand{\mknative}{\tikz[baseline=-0.6ex]\draw[mkFine,
  line width=0.8pt] (0,0) circle (0.45ex);}
\DeclareRobustCommand{\mkcoarse}{\tikz[baseline=-0.6ex]{%
  \draw[black!45,line width=1.2pt] (-0.5ex,-0.5ex) rectangle (0.5ex,0.5ex);
  \draw[mkCoarse,line width=0.7pt] (-0.5ex,-0.5ex) rectangle (0.5ex,0.5ex);}}
\DeclareRobustCommand{\mktile}{\tikz[baseline=-0.6ex]{%
  \fill[mkTileA!35] (-1.1ex,-0.55ex) rectangle (0,0.55ex);
  \fill[mkTileB!35] (0,-0.55ex) rectangle (1.1ex,0.55ex);
  \draw[white,line width=0.4pt] (0,-0.55ex) -- (0,0.55ex);}}

\title{Bifocal: Training-Free Mixed Granularity Visual Token Pruning for Multimodal LLMs}
\runningtitle{Bifocal: Mixed-Granularity Visual Token Pruning}
\author{\textbf{Peiran Yin}\textsuperscript{1,2}\quad
        \textbf{Hao Wu}\textsuperscript{1,2}\quad
        \textbf{Zihao Yang}\textsuperscript{3}\quad
        \textbf{Fengyu Zhou}\textsuperscript{1}\quad
        \textbf{Yijuan Liang}\textsuperscript{1,4}\\[2pt]
        \textbf{Haozhe Hu}\textsuperscript{1,2}\quad
        \textbf{Xudong Wang}\textsuperscript{1}\quad
        \textbf{Wei Zhang}\textsuperscript{1}\quad
        \textbf{Xiaoyu Shen}\textsuperscript{1,\textdagger}}
\affiliation{%
  \textsuperscript{1}EIT-NLP Lab, Eastern Institute of Technology, Ningbo\quad
  \textsuperscript{2}Shanghai Jiao Tong University\\
  \textsuperscript{3}Uppsala University\quad
  \textsuperscript{4}University of Science and Technology of China}
\authornotes{\textsuperscript{\textdagger}Corresponding author.}
\contact{\texttt{peiran.yin@sjtu.edu.cn}\quad\texttt{xyshen@eitech.edu.cn}}

\begin{document}
\maketitle

\begin{abstract}
Visual token pruning for multimodal large language models (MLLMs) appears highly effective, with existing methods retaining over 95\% of unpruned performance on common VQA benchmarks using a quarter of the visual tokens or fewer. However, our extensive evaluation of ten methods across three MLLM families and five ability groups reveals a critical gap: strong general performance often comes at the cost of severe degradation on fine-grained tasks such as OCR and visual grounding. We trace these failures to two limitations of the prevailing paradigm of retaining the highest-ranked native patches. A single importance ranking overlooks the complementary evidence required by different tasks, while sparse patch selection disrupts both image content and spatial integrity. To address these limitations, we propose \textbf{Bifocal}, a training-free, two-stage pruning method built on a mixed-granularity framework. It allocates separate budgets to query-focused and image-intrinsic evidence, complemented by a pooled coarse representation of the entire image. Before the LLM, Bifocal matches question words to visual regions through a soft conjunction balanced against norm-debiased image saliency, while selecting image-intrinsic evidence to maximize representational coverage. Inside the LLM, it further prunes tokens using density-adjusted text-to-image attention. With roughly a quarter of the original visual tokens, Bifocal is the only evaluated method to retain over 90\% of unpruned performance in every ability group across all three models, and ranks first on average in all nine model–budget settings.
The code is available at: \url{https://github.com/EIT-NLP/Bifocal}

\end{abstract}

\section{Introduction}
\label{sec:intro}

\begin{figure}[t]
    \centering
    \includegraphics[width=1\linewidth]{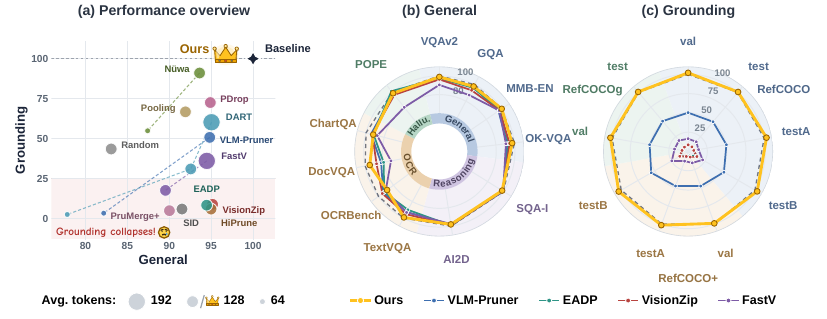}
    \vspace{-20pt}
    \caption{\textbf{Performance overview of Bifocal on LLaVA-1.5-7B.}
    \textbf{(a)}~General versus grounding performance of pruning methods in our evaluation, under matched budgets.
    \textbf{(b, c)}~Radar charts over 20 evaluations in five ability groups, at an
    average of $128$ visual tokens. Bifocal achieves state-of-the-art overall
    performance.}
    \label{fig:performance}
\end{figure}

Visual token pruning has attracted considerable attention for its ability to improve inference efficiency in multimodal large language models (MLLMs)~\citep{chen2024image,jin2025efficient,fan2025visipruner,Wu2026utp,shao2026survey,wu2026data}. Existing methods report impressive results, often retaining about $95\%$ of the unpruned performance on common VQA benchmarks with a quarter of the visual tokens or fewer. Such results have fostered a widespread view that visual tokens can already be effectively pruned with minimal performance loss. This optimism, however, does not survive a broader evaluation. We conduct an extensive evaluation of ten representative pruning methods on the widely used LLaVA-1.5-7B~\citep{liu2024improved} under three layer-averaged visual token budgets. The evaluation spans $14$ benchmarks in five ability groups: general VQA, reasoning, OCR and document understanding, object hallucination, and visual grounding.
Our evaluation reveals a starkly different picture. General VQA proves a lenient test, since even random selection or plain pooling retains most of its performance. On fine-grained tasks, by contrast, methods with near-lossless general performance degrade severely or even collapse. Figure~\ref{fig:performance}(a) illustrates this striking contrast on LLaVA-1.5-7B: with a third of the visual tokens, HiPrune~\citep{liu2026hiprune}, EADP~\citep{wang2026combating}, and SID~\citep{fang2026prune} retain $93.9\%$--$98.1\%$ of the unpruned general performance, yet only $9.5\%$--$14.0\%$ of the grounding performance.
OCR erodes as well: the OCRBench score of N\"uwa~\citep{ICLR2026_a46156bd} falls from $314$ to $152$ with $11\%$ of the tokens. Related observations have been reported in specific settings~\citep{liao2026we,peng2025can,endo2025feather,chien2025grounding,yang2026reroute}. Our evaluation further shows that this performance gap is systematic and recurs across models and token budgets, including LLaVA-NeXT and Qwen2.5-VL (Section~\ref{sec:main_results}).

We trace these failures to the paradigm shared by existing training-free methods: score each visual token by an importance estimated from the image~\citep{yang2025visionzip,zhang2025beyond,alvar2025divprune,fan2026what}, the question~\citep{pmlr-v267-zhang25s,huang2024ivtp,li2026crisp,wu2026hidrop}, or both, and keep the top-ranked native patches. Our analysis across $20$ benchmarks reveals two limitations of this paradigm.

\textbf{First, a single ranking makes the retained evidence homogeneous.} However many cues a score combines, all tokens compete in one ordering, so the budget flows to whichever kind of evidence the score favours. Yet different tasks need different evidence. Under the same budget, image-driven and question-driven selections retain largely distinct visual regions: OCR tasks favour the former, while visual grounding tasks favour the latter. Fortunately, simply splitting the budget evenly between the two already achieves performance close to the upper envelope of either selection alone across tasks.

\textbf{Second, subset-based pruning compromises the dual integrity of the visual representation.} As a subset of the original visual sequence, the retained patch set is spatially sparse and irregularly distributed, while part of the visual content is discarded. This disruption of representational integrity, rather than inaccurate token selection, limits fine-grained task performance. Even keeping exactly the patches on the ground-truth object yields only $39.1$ Acc@0.5 on RefCOCO testA. In contrast, allocating part of the budget to pooled tokens distributed across the image raises the accuracy to $63.0$, close to the unpruned $64.5$.
 
Since both limitations concern what the retained set contains, a better scoring function alone cannot remove them. We therefore ask: \textbf{what visual evidence should be preserved under a reduced token budget to support diverse tasks?} We answer with a training-free \emph{mixed-granularity} visual token pruning framework that allocates the limited budget to three complementary token sets at two granularities. For the first limitation, two fine-grained evidence sets are kept as native patches under separate budgets: \emph{query-focused evidence} preserves the details the question refers to, and \emph{image-intrinsic evidence} preserves what best represents the rest of the image. Evidence composition thus becomes an explicit allocation rather than a by-product of one ordering. For the second, a \emph{coarse context} of pooled tokens covers the whole image, summarising the discarded content and restoring an image-wide layout around the scattered fine tokens. Plugging existing methods into this framework improves their grounding accuracy by $9.7$ to $20.6$ points.


We further propose \textbf{Bifocal}, a training-free two-stage visual token pruning method. In Stage~1, before the LLM, Bifocal applies the mixed-granularity framework with dedicated selection strategies for the two fine-grained evidence sets. For query-focused evidence, we first derive an image saliency map by removing the component of CLS attention explained by patch norms. Spatially smoothed word saliency maps are then weighted by their spatial correlation with this map and combined through a weighted geometric mean, forming a soft \textsc{and} that favours patches jointly supported by informative words. For image-intrinsic evidence, we greedily select patches that maximise the marginal gain in attention-weighted representational coverage of the image, given the tokens already retained. Finally, coarse context is constructed by pooling the remaining patches within spatial blocks, completing the mixed-granularity representation. In Stage~2, inside the LLM, Bifocal further removes redundant visual tokens using text-to-image attention adjusted by local token density.

In our evaluation, Bifocal is the only method that consistently retains over $90\%$ of the unpruned performance across all five ability groups on all three models with roughly a quarter of the visual tokens. In contrast, every baseline falls below $90\%$ in at least one group, typically grounding or OCR. On LLaVA-1.5-7B with a layer-averaged budget of $128$ visual tokens, Bifocal's weakest group, OCR, still retains $92.4\%$, while the second-best method on grounding retains only $86.3\%$ on OCR. On Qwen2.5-VL-7B with $25\%$ of the visual tokens, the strongest baseline retains $89.2\%$ on OCR and $85.4\%$ on grounding, compared with $92.3\%$ and $90.2\%$ for Bifocal. On LLaVA-NeXT-7B with a layer-averaged budget of approximately $640$ visual tokens, no baseline retains more than $73\%$ of the unpruned grounding performance, whereas Bifocal retains $90.7\%$. Across all nine model--budget settings, Bifocal also ranks first in both general and grounding averages.

Our contributions are threefold.
\textbf{(1)~Diagnosis.} Motivated by the lopsided performance of existing pruners, we examine $20$ benchmarks under a small token budget. First, image- and question-driven visual evidence are complementary across tasks, and explicitly allocating the budget between them improves robustness. Second, mean pooling neatly restores both the content and positional integrity broken by pruning.
\textbf{(2)~Framework.} We propose a mixed-granularity framework that combines separately budgeted query-focused and image-intrinsic evidence with pooled coarse context, and that also improves existing methods.
\textbf{(3)~Method.} We propose Bifocal, which instantiates the framework with dedicated selection rules. Bifocal achieves state-of-the-art averages in all nine model--budget settings and, with roughly a quarter of the visual tokens, retains over $90\%$ of the unpruned performance in every ability group.

\section{Related Work}
\label{app:related}
\paragraph{Text-Guided Visual Token Pruning.}

Within an LLM, textual hidden states progressively integrate linguistic context and visual information through multiple layers of causal self-attention. A simple and computationally efficient strategy is therefore to use the final text token as a proxy for the full instruction and estimate visual token importance from its attention to visual tokens~\citep{chen2024image,xing2025conical,endo2025feather}. Building on this idea, subsequent methods select or aggregate multiple text tokens to extract task-relevant textual cues for visual token pruning~\citep{pmlr-v267-zhang25s,Singh_2026_CVPR,liu2024multi}. Beyond directly reusing existing textual hidden states, some studies explicitly construct or learn text-conditioned representations for visual token pruning. IVTP~\citep{huang2024ivtp} constructs a pseudo-CLS token that is semantically associated with the current instruction and uses it as a reference for visual token selection. LVPruning~\citep{sun2025lvpruning} introduces learnable cross-attention modules to explicitly fuse textual and visual representations and predict visual token retention decisions.

Before visual features enter the LLM, instruction embeddings have not undergone deep contextualization or extensive interaction with the current visual input, although pretrained vision encoders and modality projectors already provide a certain degree of cross-modal semantic alignment. Accordingly, methods such as TRIM~\citep{song2025less} exploit embedding-level text--vision relevance by measuring the similarity between CLIP text embeddings and visual patch embeddings, while subsequent approaches apply pooling or aggregation to derive text--vision relevance scores~\citep{zhang2025zspaprune,han2025adav,liu2026prompt}. Rather than treating the entire instruction as a single unit, CRISP~\citep{li2026crisp} extracts keywords and compares them with projected visual tokens, whereas ICCTP~\citep{yu2026instruction} uses instruction nouns as visual-semantic anchors for cross-modal clustering and visual token compression.

\paragraph{Image-Guided Visual Token Pruning.}
Unlike methods conditioned on the current textual instruction, image-guided approaches compress visual tokens solely according to intrinsic visual cues, such as saliency, informativeness, similarity, and redundancy. One line of work estimates visual token saliency from the attention maps produced by the vision encoder. A common strategy is to use the attention from the CLS token to image patches to identify visually important tokens~\citep{shang2025llava}. HiPrune~\citep{liu2026hiprune} further exploits distinct attention patterns across different vision-encoder layers to capture object-level and global visual information, while EntropyPrune~\citep{wang2026entropyprune} measures visual token informativeness from an information-theoretic perspective using representation entropy. Another line of work focuses on similarity and redundancy among visual tokens, reducing repeated information by retaining representative or diverse visual features~\citep{yang2025visionzip,alvar2025divprune}. VisPruner~\citep{zhang2025beyond} combines these two types of visual cues by first selecting salient tokens according to visual attention and then removing redundant tokens based on feature similarity.

\paragraph{Joint Text--Image-Guided Visual Token Pruning.}
Some approaches jointly exploit textual and visual cues for visual token compression. Multi-stage pruning strategies naturally fall into this category because they use complementary guidance signals at different compression stages. MustDrop~\citep{liu2024multi} first removes redundant tokens according to visual similarity during visual encoding and subsequently leverages textual semantics to further filter visual information. DUET-VLM~\citep{Singh_2026_CVPR} first performs redundancy-aware merging based on correlations among visual tokens and then applies progressive pruning within the LLM using text-to-vision attention.
Another line of work combines textual and visual signals within a unified selection process to estimate visual token importance. AdaV~\citep{han2025adav} and CrisPrune~\citep{liu2026crisprune} both use ViT CLS--patch self-attention to characterize visual saliency and the cosine similarity between visual tokens and text embeddings to estimate textual relevance. AdaV aggregates textual relevance using the maximum similarity, whereas CrisPrune further combines the two signals into a token-quality score. CoIn~\citep{du2026coin} measures intrinsic visual saliency using the $\ell_2$ norm of visual token embeddings and evaluates textual relevance through the cosine similarity between visual tokens and the mean-pooled text embedding. The two normalized scores are then combined to obtain an overall informativeness score.

\definecolor{routeImg}{HTML}{E69F00}
\definecolor{routeTxt}{HTML}{0072B2}
\definecolor{routeBoth}{HTML}{009E73}
\definecolor{routeRand}{HTML}{999999}
\providecommand{\sw}[1]{{\color{#1}\rule[-0.15ex]{1.3ex}{1.3ex}}}

\section{Mixed-Granularity Pruning Framework}
\label{sec:framework}

\subsection{Information: Two Sources}
\label{sec:sources}

Existing training-free pruning methods often exhibit pronounced task-wise biases ~\citep{endo2025feather,chien2025grounding}. We therefore turn the question around: \keyq{what visual evidence does an LLM actually need to solve different tasks?}

At inference time, a training-free pruning method has access to only two sources of information: the image itself and the question. We therefore examine how token selections driven by each source behave under the same budget. We fix the token budget at one quarter of the visual tokens and turn each source into a route with a common, simple scoring rule: the image route ranks patches by CLS attention, and the text route ranks them by their cosine similarity to the mean word embedding of the question.

\begin{figure}[htbp]
\centering
\includegraphics[width=\linewidth]{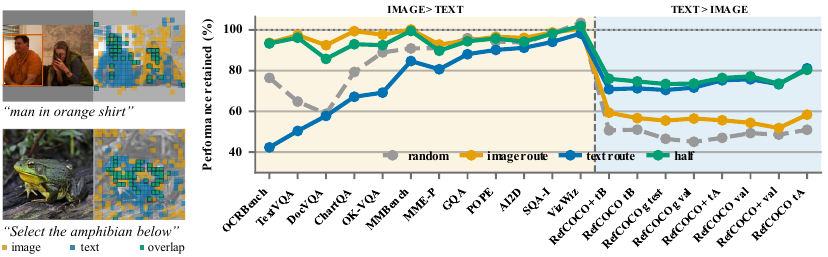}
\caption{
\textbf{The image and text routes under a fixed token budget.}
\sw{routeImg}~image route; \sw{routeTxt}~text route.
\textbf{Left:} tokens selected by each route on two examples;
\sw{routeBoth}~marks tokens selected by both.
\textbf{Right:} performance relative to the baseline on $20$ benchmarks at $144$ of $576$ tokens;
\sw{routeBoth}~\emph{half} splits the budget evenly between the two routes, and
\sw{routeRand}~\emph{random} retains tokens at random.
}
\label{fig:routes}
\end{figure}

Figure~\ref{fig:routes} (left) shows that the two routes retain noticeably different regions, and their selections overlap only slightly. Figure~\ref{fig:routes} (right) shows that this difference is systematic across tasks: OCR and document tasks favour the image route, REC tasks favour the text route, and the general benchmarks show much smaller gaps. Relying on either source alone therefore leads to a clear task-wise bias, which suggests that the two sources provide complementary rather than interchangeable evidence. We further include two references: random retention and an even split of the budget between the two routes. The even-split curve largely tracks the upper envelope of the two single-route curves; its advantage is not concentrated on any particular task but lies in robustness across tasks. \apx{Appendix~\ref{app:single} analyses the effect of each single route.}

\subsection{Structure: Two Integrities}
\label{sec:integrity}

\begin{figure}[ht]
  \centering
    \begin{minipage}[t]{0.59\linewidth}
  \vspace{0pt}
  \centering
  \footnotesize
  \renewcommand{\arraystretch}{1.3}%
  \setlength{\tabcolsep}{3.5pt}%
  \begin{tabular*}{\linewidth}{@{}cl@{\extracolsep{\fill}}cc@{}}
    \toprule
    & Visual tokens & Center $\uparrow$ & ACC@0.5 $\uparrow$ \\
    \midrule
    \cnum{1} & No pruning, $576$ patches     & 91.07 / 85.16 & 64.47 / 47.48 \\
    \cnum{2} & $144$ patches, both routes    & 85.95 / 77.64 & 51.83 / 35.47 \\
    \cnum{3} & $144$ patches on the GT bbox  & 71.92 / 65.78 & 39.14 / 27.80 \\
    \addlinespace
    \cnum{4} & $80$ patches $+$ $64$ rim     & 81.37 / 79.25 & 42.50 / 29.62 \\
    \cnum{5} & $80$ patches $+$ $64$ random  & 88.19 / 84.20 & 54.52 / 40.82 \\
    \cnum{6} & $80$ patches $+$ $64$ lattice & 87.64 / 84.79 & 54.98 / 40.88 \\
    \addlinespace
    \cnum{7} & $144$ summaries, no patches   & 85.66 / 77.74 & 44.62 / 29.11 \\
    \cnum{8} & $80$ patches $+$ $64$ blanked & \underline{89.92} / \textbf{86.65} & \underline{56.62} / \underline{42.59} \\
    \rowcolor{oursbg}
    \cnum{9} & $80$ patches $+$ $64$ pooled  & \textbf{90.53} / \underline{86.12} & \textbf{63.02} / \textbf{46.18} \\
    \bottomrule
  \end{tabular*}
\end{minipage}%
  \hfill
  \begin{minipage}[t]{0.38\linewidth}
    \vspace{0pt}
    \centering
    \includegraphics[width=\linewidth]{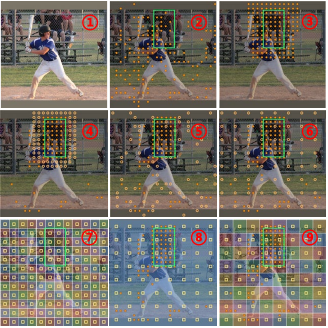}
  \end{minipage}
  \caption{
  \textbf{Left:} experiments testing positional and content integrity. Every
  row except \cnum{1} keeps $144$ of the $576$ visual tokens; each cell
  reports RefCOCO testA / testB, with the best pruned result in bold and the
  second best underlined.
  \textbf{Right:} the tokens each row retains on one example.
  \mkgt~ground-truth box; \mkpred~predicted box; \mkfine~retained patch;
  \mknative~extra native patch; \mkcoarse~pooled token, averaging its
  \mktile~tile.}
  \label{fig:coarse}
  \vspace{-10pt}
\end{figure}

Even after combining the two routes, grounding remains the least recovered
capability. Is the remaining gap simply a matter of inaccurate token
selection? Figure~\ref{fig:coarse} suggests otherwise. In \cnum{3}, we push
selection to the extreme and retain only the patches on the ground-truth
target, yet localisation falls well below the two-route selection in
\cnum{2}. Conventional visual token pruning methods, including our two routes, share
the same compression paradigm: retaining a subset of native patches from the
full visual sequence. \keyq{This subset-based compression breaks the integrity
of the original visual sequence in two ways: positional integrity and content
integrity.}

\paragraph{Positional integrity.}
Subset pruning leaves tokens at only a sparse and irregular set of positions,
breaking the global positional organization of the original visual sequence.
Starting from this extreme strategy in \cnum{3}, we progressively restore coverage beyond the target---first around its boundary (\cnum{4}), then across the full image with random or lattice layouts (\cnum{5}, \cnum{6}). Localisation improves substantially as the
retained tokens recover a more complete image-wide distribution.

\paragraph{Content integrity.}
Token pruning also incurs an unavoidable loss of content. To examine this
effect, we keep the image-wide positional layout fixed and vary only what the
distributed tokens contain. In \cnum{8}, all distributed tokens carry the
same image-wide average, yet localisation improves over the native lattice in
\cnum{6}. Replacing this global summary with region-specific summaries
(\cnum{9}) improves performance further. Discarded
content need not be preserved at its original resolution, but retaining a
low-resolution account of it is important for content integrity.

A layer of pooled coarse tokens addresses both deficiencies at once: each
coarse token summarizes content that would otherwise be discarded, while
together they restore an even positional layout over the image. The strong
performance of \cnum{9} confirms the effectiveness of this design. This
closely mirrors the coarse--fine decomposition of classical image
pyramids~\citep{burt1987laplacian}, where local detail and global structure are represented at
different resolutions. It also motivates our mixed-granularity design: fine
tokens preserve task-relevant native detail, while coarse tokens recover, at
a lower resolution, the content and positional integrity lost through subset
selection. \apx{Appendix~\ref{app:plug} validates the framework by integrating existing pruning methods into it.}


\section{Methodology}
\label{sec:method}

\begin{figure}[t]
  \centering
  \includegraphics[width=\linewidth]{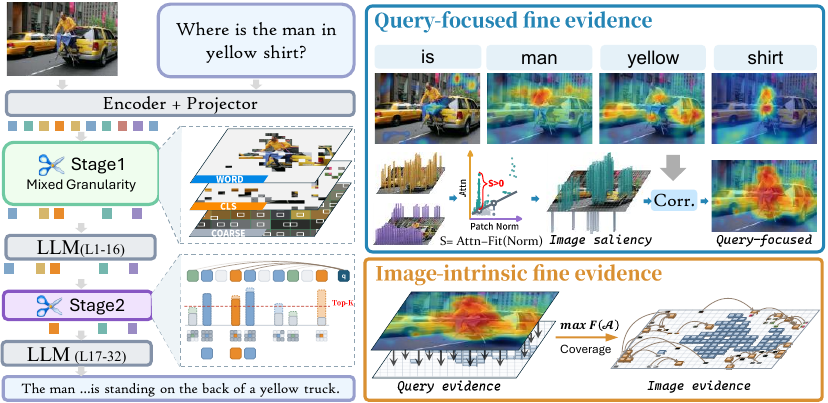}
    \caption{\textbf{Overview of Bifocal}, 
    a training-free, two-stage visual token pruning method. 
    \textbf{Stage~1} runs before the LLM and retains a
mixed-granularity representation: fine evidence at native-patch resolution,
of two kinds, query-focused evidence selected for the question and
image-intrinsic evidence selected from the image itself, and coarse tokens obtained by average pooling. 
    \textbf{Stage~2} keeps the top-$K$ tokens by text-to-image attention inside the LLM.}
  \label{fig:overview}
\end{figure}

Figure~\ref{fig:overview} gives an overview of the design of Bifocal. We
describe query-focused evidence in \S\ref{sec:query}, image-intrinsic evidence
in \S\ref{sec:intrinsic}, coarse context in \S\ref{sec:coarse}, and Stage~2
in \S\ref{sec:stage2}.

\subsection{Query-Focused Evidence}
\label{sec:query}

Query-focused evidence assigns each patch a question-dependent score $I_j$ and
retains the top-$K$ patches as the evidence the question needs. Existing
methods typically compress the question into a single vector and score each
patch by its cosine similarity to it. Whether the words are pooled by a max or
a mean, a strong match to one word can stand in for the absence of another,
an \textsc{or}-like rule under which a ``red shirt'' is no different from a red fire
hydrant elsewhere in the image.
Since no model reads the question as a sentence before the LLM, we treat it as
a set of words and model two ingredients explicitly: word weighting and
word conjunction. For weighting, we invite the image back to sit on the
jury alongside the question: a word earns a louder voice when its saliency map
agrees with what the image itself finds salient. For conjunction, we combine the words with a geometric
mean, a soft \textsc{and} that favours patches supported by all relevant words jointly.

\paragraph{Word Saliency Maps}
\label{sec:word}



We construct a word saliency map for each word to locate the image patches it attends to.
Let $e_w$ be the embedding of word $w$ and $p_j$ the feature of patch $j$ after the projector maps it into the LLM space. Cosine similarity measures how strongly word $w$ attends to patch $j$:
\begingroup
\small
\setlength{\abovedisplayskip}{3pt}
\setlength{\belowdisplayskip}{3pt}
\setlength{\abovedisplayshortskip}{3pt}
\setlength{\belowdisplayshortskip}{3pt}
\begin{equation}
s_w(j)=\cos(\hat e_w,\ \hat p_j),\qquad \hat e_w=e_w/\|e_w\|.
\end{equation}
\endgroup
In fact, we observe that this commonly used operation of assigning each patch a score via cosine similarity leads to large score gaps between neighboring patches on the same object. Such discrepancies arise from estimation noise rather than genuine differences in content. Because patches are selected by ranking these scores, the noise can promote patches that happen to score high while displacing those that actually lie on the target. Fortunately, signal and noise occupy different spatial scales: an object spans several patches, so the correlation of interest is spatially coherent, whereas estimation noise fluctuates independently from patch to patch. We find that, on a word saliency map, a simple spatial smoothing $\mathcal{G}$ is enough: this low-pass filter averages away uncorrelated noise while leaving extended structure largely intact, yielding $\tilde s_w=\mathcal{G}\,s_w$. We then min-max normalise each smoothed map over the patches, $\hat s_w=\mathrm{mm}(\tilde s_w)$, so that the conjunction compares the spatial shape of the word maps rather than their overall level, which is typically high for content words and low for function words.
\apx{Appendix~\ref{app:smooth} analyses the smoothing further.}

\paragraph{Image Saliency Map}
\label{sec:visual}

Word weighting needs a referee for the word saliency maps, and \apx{Appendix~\ref{app:align}} suggests a natural one: what the image itself attends to after the ViT. Following common practice, we start from the attention of the ViT's CLS token, which aggregates global information. It is not, however, a clean content map: a substantial part of the CLS score follows the norm of a patch rather than its content, most visibly in the attention sink~\citep{darcet2024vision,xiao2024efficient}, a few patches of abnormally large norm that mostly sit on low-information background. Since this map weights the words, one that points at the background hands the say to words unrelated to the query.

The reason, derived and verified in \apx{Appendix~\ref{app:norm}}, is that the
CLS attention factorises exactly as $a_j=n_j\,c_j$, a gain $n_j$ set by the
norm times a content score $c_j$. To obtain the content saliency, we need to
decouple the norm from the content. Neither factor is observed, so we separate
the product directly in the attention, using only the attention scores and the
norm. By the second-order Taylor expansion
of $xy$ about $(\bar x,\bar y)$, which is exact, for any $x,y$ and any
reference values $\bar x,\bar y$,
\begingroup
\small
\setlength{\abovedisplayskip}{3pt}
\setlength{\belowdisplayskip}{3pt}
\setlength{\abovedisplayshortskip}{3pt}
\setlength{\belowdisplayshortskip}{3pt}
\begin{equation}
xy=\bar y\,x+\bar x\,y-\bar x\bar y+(x-\bar x)(y-\bar y).
\label{eq:identity}
\end{equation}
\endgroup
Substituting $x=n_j$ and $y=c_j$, with the reference values taken to be their
image-wide means $\bar n$ and $\bar c$:
\begingroup
\small
\setlength{\abovedisplayskip}{3pt}
\setlength{\belowdisplayskip}{3pt}
\setlength{\abovedisplayshortskip}{3pt}
\setlength{\belowdisplayshortskip}{3pt}
\begin{equation}
a_j=\bar c\,n_j\;+\;\bar n\,c_j\;-\;\bar n\bar c\;+\;\big(n_j-\bar n\big)\big(c_j-\bar c\big).
\label{eq:decomp}
\end{equation}
\endgroup
With the means as reference values, the four terms separate cleanly. The first
depends only on the gain and is linear in it; the second ranks patches exactly
as $c_j$ does; the third is a constant; only the cross term involves both, and
it is small wherever the gain stays near its mean. Removing the first term
therefore leaves, up to a constant and the cross term, a map that ranks patches
by content. The first term is not observable, but the gain rises with the norm,
so over one image we approximate it by a linear function of the smoothed norm
map $\nu$ of the projected patch features: we regress the smoothed CLS attention
$\tilde a=\mathcal G a$ on $\nu$ over the patches of the image by least squares
and subtract the fit patch by patch. If content does not co-vary with the norm,
the fit recovers the first term; whatever does co-vary with the norm is removed
along with it. The residual is the image saliency map:
\begingroup
\small
\setlength{\abovedisplayskip}{3pt}
\setlength{\belowdisplayskip}{3pt}
\setlength{\abovedisplayshortskip}{3pt}
\setlength{\belowdisplayshortskip}{3pt}
\begin{equation}
S_j=\tilde a_j-\big(\hat\beta\,\nu_j+\hat b\big).
\label{eq:S}
\end{equation}
\endgroup

\paragraph{Word Weighting and Word Conjunction}
\label{sec:weighting}

With both maps in hand, each word is weighted by how well its saliency map
agrees with the image saliency map:
\begingroup
\small
\setlength{\abovedisplayskip}{3pt}
\setlength{\belowdisplayskip}{3pt}
\setlength{\abovedisplayshortskip}{3pt}
\setlength{\belowdisplayshortskip}{3pt}
\begin{equation}
\alpha_w \;\propto\; \mathrm{relu}\big(\mathrm{corr}(\hat s_w,\ S)\big), \qquad \sum_w \alpha_w = 1,
\label{eq:alpha}
\end{equation}
\endgroup
where $\mathrm{corr}$ is the Pearson correlation over patches and
$\mathrm{relu}$ discards words whose map disagrees with the image. The words are
then combined by a weighted geometric mean, a soft AND under which a patch
scores high only if every weighted word points to it:
\begingroup
\small
\setlength{\abovedisplayskip}{3pt}
\setlength{\belowdisplayskip}{3pt}
\setlength{\abovedisplayshortskip}{3pt}
\setlength{\belowdisplayshortskip}{3pt}
\begin{equation}
R_j \;=\; \prod_w \hat s_w(j)^{\,\alpha_w}.
\label{eq:R}
\end{equation}
\endgroup
The norm map $\nu$ removed above also carries information of its own:
high-norm patches mostly lie on background, so $B=-\nu$ serves as a
query-independent background penalty. The query-focused score combines the
three maps by an unweighted geometric mean of their min-max normalised versions:
\begingroup
\small
\setlength{\abovedisplayskip}{3pt}
\setlength{\belowdisplayskip}{3pt}
\setlength{\abovedisplayshortskip}{3pt}
\setlength{\belowdisplayshortskip}{3pt}
\begin{equation}
I_j \;=\; \big[\mathrm{mm}(R_j)\cdot \mathrm{mm}(S_j)\cdot \mathrm{mm}(B_j)\big]^{1/3}
\label{eq:I}
\end{equation}
\endgroup

\subsection{Image-Intrinsic Evidence}
\label{sec:intrinsic}

With part of the token budget spent on query-focused evidence, and following
the analysis of Figure~\ref{fig:routes}, we now select a second kind of evidence
that comes from the image itself. Query-focused evidence concentrates on the
object the question names; image-intrinsic evidence spends its budget elsewhere,
on the $N_i$ patches that best represent the rest of the image. This calls for a
measure of how well the image is represented. Let $\mathcal A$ be the set of
tokens currently in place. For each patch $j$ we take its similarity to the
closest token in $\mathcal A$, and sum these values weighted by the CLS
attention $a_j$. We call the result the \emph{representational coverage} of
$\mathcal A$:
\begingroup
\small
\setlength{\abovedisplayskip}{3pt}
\setlength{\belowdisplayskip}{3pt}
\setlength{\abovedisplayshortskip}{3pt}
\setlength{\belowdisplayshortskip}{3pt}
\begin{equation}
F(\mathcal A) \;=\; \sum_j a_j \cdot
\max\left(0,\max_{i \in \mathcal A} \cos(x_j,\, x_i)\right)
\label{eq:cover}
\end{equation}
\endgroup
Because of the $\max$, a patch needs only one representative, and a near-copy
of a token already in place adds nothing. $F$ is monotone and submodular, so
greedy selection is near-optimal: at each step we add the patch with the
largest marginal gain,
\begingroup
\small
\setlength{\abovedisplayskip}{3pt}
\setlength{\belowdisplayskip}{3pt}
\setlength{\abovedisplayshortskip}{3pt}
\setlength{\belowdisplayshortskip}{3pt}
\begin{equation}
\Delta(t)\;=\;F(\mathcal A\cup\{t\})-F(\mathcal A)
\;=\;\sum_j a_j\,\mathrm{relu}\big(\cos(x_t,x_j)-\mathrm{cov}_j\big),
\qquad
\mathrm{cov}_j=\max\left(0,\max_{i\in\mathcal A}\cos(x_j,x_i)\right)
\label{eq:gain}
\end{equation}
\endgroup
where $\mathrm{cov}_j$ is how well patch $j$ is represented so far
(Algorithm~\ref{alg:cover}, Appendix~\ref{app:cover}).

\subsection{Coarse Context}
\label{sec:coarse}

Following the pooling analysis of Figure~\ref{fig:coarse}, we add a layer of pooled coarse tokens after the fine evidence. We make one adaptation here: the content of each pooling block averages only the patches inside the block that were \emph{not} selected as fine evidence, and its position is taken to be the attention-weighted centroid of the block. Let $\mathcal B_b$ denote the members of block $b$ and $\mathcal Q$ the patches selected as fine evidence, so that the residual set is $\mathcal R_b=\mathcal B_b\setminus\mathcal Q$. Then
\begingroup
\small
\setlength{\abovedisplayskip}{3pt}
\setlength{\belowdisplayskip}{3pt}
\setlength{\abovedisplayshortskip}{3pt}
\setlength{\belowdisplayshortskip}{3pt}
\begin{equation}
\bar x_b=\sum_{j\in\mathcal R_b}\frac{a_j}{\sum_{k\in\mathcal R_b}a_k}\,x_j,
\qquad
\hat r_b=\Big\lfloor\sum_{j\in\mathcal B_b}\frac{a_j}{\sum_{k\in\mathcal B_b}a_k}\,r_j+\tfrac{1}{2}\Big\rfloor,
\label{eq:coarse}
\end{equation}
\endgroup
where $r_j$ is the grid coordinate of patch $j$, the rounding puts the centroid back on the nearest patch inside the block, and $\hat r_b$ is the position this coarse token takes in the sequence. Under a fixed budget this reduces redundancy among the tokens: every patch is either present natively or enters exactly one coarse token.

\subsection{Stage~2: Pruning inside the LLM}
\label{sec:stage2}

Stage~1 decides which visual evidence enters the LLM, yet the visual tokens
remain highly redundant inside it~\citep{chen2024image}. By then the causal
attention has also let them absorb the question, so the text-to-image attention
reliably reflects which visual information the answer still needs. Following
FastV, at layer $L$ we score each visual token by the attention $t_j$ it
receives from the last text token and keep the top-$K$.

We make one adjustment, with the integrity of the retained set in mind. After
Stage~1 the grid is sparse, and this attention often peaks on isolated
survivors whose neighbours have been pruned. We therefore weight each score by
a \emph{local-density prior}:
\begingroup
\small
\setlength{\abovedisplayskip}{3pt}
\setlength{\belowdisplayskip}{3pt}
\setlength{\abovedisplayshortskip}{3pt}
\setlength{\belowdisplayshortskip}{3pt}
\begin{equation}
\tilde t_j=t_j\cdot d_j,
\label{eq:s2}
\end{equation}
\endgroup
where $d_j$ is the fraction of the neighbourhood of token $j$ that survived
Stage~1. Pruned patches no longer receive attention; they enter Stage~2 only by
lowering the density of their neighbours.

\vspace{-10pt}
\section{Experiments}
\vspace{-5pt}

We evaluate Bifocal at three budgets on LLaVA-1.5-7B
(Tables~\ref{tab:general} and~\ref{tab:grounding}), Qwen2.5-VL-7B
(Table~\ref{tab:qwen}) and LLaVA-NeXT-7B (\apx{Appendix~\ref{app:next}}),
followed by ablations (\S\ref{sec:ablation}), an efficiency analysis
(\apx{Appendix~\ref{app:efficiency}}) and visualisations
(\S\ref{sec:visual-study}).

\begin{table}[t]
\centering
\caption{\textbf{General multimodal benchmarks on LLaVA-1.5-7B.} Methods are
matched by layer-averaged visual-token budget. Within each budget, the best and
second-best results are \textbf{bold} and \underline{underlined}. Avg.\ averages
benchmark-wise performance relative to the baseline. }
\label{tab:general}
\scriptsize
\setlength{\tabcolsep}{1.4pt}\renewcommand{\arraystretch}{1.05}
\setlength{\aboverulesep}{0pt}\setlength{\belowrulesep}{0pt}
\begin{tabularx}{\linewidth}{@{}l *{11}{>{\centering\arraybackslash}X} >{\columncolor{black!4}\centering\arraybackslash}X@{}}
\toprule
\multirow{2}{*}{Method} & \cat{4}{catGen}{General} & \cat{2}{catRea}{Reasoning} & \cat{4}{catOcr}{OCR} & \cat{1}{catHal}{Hall.} & \cat{1}{black}{Avg.} \\
\cmidrule(lr){2-5}\cmidrule(lr){6-7}\cmidrule(lr){8-11}\cmidrule(lr){12-12}
 & VQA\textsuperscript{v2} & GQA & MMB & VQA\textsuperscript{OK} & SQA-I & AI2D & VQA\textsuperscript{T} & OCR-B & DocVQA & ChartQA & POPE & \\
\midrule
\ub{Baseline} & \ub{78.5} & \ub{61.9} & \ub{64.1} & \ub{53.5} & \ub{69.4} & \ub{55.2} & \ub{46.0} & \ub{314} & \ub{21.5} & \ub{18.2} & \ub{85.9} & \ub{100.0\%} \\
\midrule

\rowcolor{tierbg}\multicolumn{13}{c}{\textbf{Average Tokens = 192}} \\
FastV & 75.0 & 57.0 & 63.6 & 51.1 & \underline{69.1} & 54.1 & 44.6 & 300 & 18.2 & 17.1 & 75.8 & 94.5\% \\
PDrop & 77.4 & 60.0 & 63.5 & 52.4 & 68.9 & 54.1 & \underline{45.9} & 308 & 19.9 & \textbf{18.2} & 85.3 & \underline{98.1\%} \\
PruMerge+ & 74.6 & 57.6 & 61.3 & 48.6 & 69.0 & 54.6 & 39.0 & 282 & 14.8 & 16.5 & 79.9 & 90.9\% \\
VisionZip & 76.8 & 59.3 & \underline{63.8} & 52.8 & 68.7 & \underline{55.5} & 44.5 & \underline{312} & 19.4 & 17.4 & 85.3 & 97.5\% \\
SID & 76.4 & 58.9 & 63.3 & 51.9 & 68.9 & 54.9 & 40.9 & 289 & 15.4 & 17.0 & 85.8 & 93.9\% \\
DART & 76.7 & 59.5 & 62.9 & 51.3 & 68.9 & 54.7 & 43.6 & 300 & 19.6 & 17.5 & 81.8 & 96.2\% \\
N\"uwa & 77.8 & 61.0 & \underline{63.8} & \textbf{53.5} & 68.6 & 54.4 & 45.7 & 244 & 20.2 & \textbf{18.2} & 84.6 & 96.7\% \\
EADP & 77.2 & 60.3 & 62.9 & 51.3 & 68.8 & 54.5 & 43.3 & 301 & 19.3 & 17.2 & \textbf{87.4} & 96.7\% \\
VLM-Pruner & \underline{77.9} & \underline{61.2} & 62.0 & 51.3 & \textbf{69.9} & 55.1 & 44.1 & 299 & 19.5 & 17.6 & 85.6 & 97.2\% \\
HiPrune & 76.8 & 59.2 & 63.0 & 52.4 & 68.8 & 54.5 & 44.9 & \textbf{316} & \underline{20.5} & \underline{17.8} & \underline{86.1} & \underline{98.1\%} \\

\rowcolor{oursbg} \textbf{Bifocal (ours)} & \textbf{78.2} & \textbf{61.5} & \textbf{64.4} & \underline{53.4} & 68.4 & \textbf{55.6} & \textbf{46.2} & 284 & \textbf{21.6} & 17.0 & 85.7 & \textbf{98.5\%} \\

\rowcolor{tierbg}\multicolumn{13}{c}{\textbf{Average Tokens = 128}} \\
FastV & 71.0 & 54.1 & 62.0 & 48.6 & 69.1 & 53.9 & 42.9 & 284 & 15.4 & 16.0 & 68.2 & 89.5\% \\
PDrop & 74.9 & 57.4 & 61.4 & 50.4 & \textbf{70.9} & 52.9 & \underline{44.5} & \textbf{299} & 18.2 & \underline{17.5} & 81.8 & 94.9\% \\
PruMerge+ & 74.0 & 57.6 & 60.3 & 47.2 & \underline{69.7} & 54.3 & 39.2 & 279 & 14.2 & 15.9 & 81.1 & 90.0\% \\
VisionZip & 75.6 & 57.7 & 62.2 & 50.7 & 68.8 & 54.7 & 43.8 & \underline{298} & 18.8 & 17.0 & 82.9 & \underline{95.2\%} \\
SID & 75.4 & 58.0 & 62.2 & 50.1 & 68.5 & 54.6 & 39.3 & 272 & 14.7 & 16.4 & 84.4 & 91.5\% \\
DART & 74.8 & 57.8 & 61.6 & 49.9 & 69.6 & 53.9 & 41.6 & 275 & 17.0 & 17.2 & 77.8 & 92.6\% \\
N\"uwa & \underline{77.1} & \underline{60.3} & \underline{63.1} & \underline{51.9} & 68.2 & 54.2 & 43.5 & 202 & 19.0 & \textbf{17.8} & 83.6 & 93.6\% \\
EADP & 76.5 & 59.6 & 62.7 & 50.1 & 68.8 & \underline{54.8} & 41.4 & 290 & 17.2 & 16.7 & \textbf{87.2} & 94.5\% \\
VLM-Pruner & \underline{77.1} & \underline{60.3} & 61.4 & 49.3 & 69.3 & 54.2 & 42.6 & 294 & 17.8 & 17.2 & 84.4 & 94.8\% \\
HiPrune & 74.9 & 57.3 & 61.9 & 50.5 & 68.4 & \textbf{55.2} & 43.6 & 296 & \underline{19.3} & 16.8 & 82.8 & 95.0\% \\
\rowcolor{oursbg} \textbf{Bifocal (ours)} & \textbf{77.6} & \textbf{60.5} & \textbf{64.1} & \textbf{52.1} & 68.2 & 54.2 & \textbf{45.6} & 262 & \textbf{20.4} & 16.8 & \underline{84.7} & \textbf{96.3\%} \\

\rowcolor{tierbg}\multicolumn{13}{c}{\textbf{Average Tokens = 64}} \\
SID & 74.2 & 57.3 & 59.5 & \underline{48.4} & 68.3 & \underline{53.6} & 38.0 & \textbf{276} & 13.8 & \underline{16.2} & \underline{84.3} & 89.8\% \\
DART & 65.4 & 51.1 & 54.8 & 42.5 & 68.8 & 51.7 & 29.3 & 214 & 11.7 & 14.2 & 59.0 & 77.8\% \\
N\"uwa & \underline{74.6} & \textbf{58.6} & \underline{62.0} & \textbf{49.4} & 67.8 & 53.0 & \underline{39.9} & 152 & 15.5 & \textbf{16.3} & 79.8 & 87.4\% \\
EADP & \textbf{75.3} & 58.1 & 60.7 & 47.9 & \underline{69.1} & \textbf{53.9} & 37.6 & \underline{268} & 14.4 & 15.6 & \textbf{85.6} & \underline{90.0\%} \\
VLM-Pruner & 69.9 & 54.4 & 57.7 & 43.0 & \textbf{69.3} & 52.5 & 32.8 & 240 & 11.9 & 14.3 & 69.1 & 82.2\% \\
HiPrune & 69.1 & 53.6 & 59.5 & 46.3 & 68.9 & 53.4 & \underline{39.9} & 258 & \underline{16.1} & 16.1 & 73.0 & 87.9\% \\
\rowcolor{oursbg} \textbf{Bifocal (ours)} & \textbf{75.3} & \underline{58.5} & \textbf{62.7} & 46.8 & 67.6 & 52.5 & \textbf{40.9} & 254 & \textbf{17.0} & 15.4 & 81.7 & \textbf{90.6\%} \\
\bottomrule
\end{tabularx}
\end{table}

\subsection{Experimental Setup}
\vspace{-5pt}
\paragraph{Models and benchmarks.}

We report main results on LLaVA-1.5-7B and verify on LLaVA-NeXT-7B and
Qwen2.5-VL-7B. The benchmarks are
VQA\textsuperscript{v2}~\citep{goyal2017making},
GQA~\citep{hudson2019gqa},
MMBench~\citep{liu2024mmbench},
OK-VQA~\citep{marino2019ok},
ScienceQA-IMG~\citep{lu2022learn},
AI2D~\citep{kembhavi2016diagram},
TextVQA~\citep{singh2019towards},
OCRBench~\citep{liu2024ocrbench},
DocVQA~\citep{mathew2021docvqa},
ChartQA~\citep{masry2022chartqa},
POPE~\citep{li2023evaluating},
RefCOCO, RefCOCO+~\citep{yu2016modeling} and
RefCOCOg~\citep{mao2016generation}.

\paragraph{Baselines.}
We select representative visual-token pruning methods as baselines, including
FastV~\citep{chen2024image}, PDrop~\citep{xing2025conical},
VisionZip~\citep{yang2025visionzip}, and PruMerge+~\citep{shang2025llava}.
We further reproduce recent publicly available representative works and conduct
fair comparisons with open-source methods that provide favorable reproducibility,
including DART~\citep{wen2025stop}, N\"uwa~\citep{ICLR2026_a46156bd},
HiPrune~\citep{liu2026hiprune}, EADP~\citep{wang2026combating},
SID~\citep{fang2026prune}, and VLM-Pruner~\citep{wu2026vlm}.

\subsection{Main Results}
\label{sec:main_results}

\begin{table}[ht]
\centering
\caption{\textbf{Visual grounding on LLaVA-1.5-7B.} All entries are
\textsc{Acc}@$0.5$. Budget groups and highlighting follow
Table~\ref{tab:general}.}
\label{tab:grounding}
\scriptsize
\setlength{\tabcolsep}{2.5pt}\renewcommand{\arraystretch}{0.9}
\setlength{\aboverulesep}{0pt}\setlength{\belowrulesep}{0pt}
\begin{tabularx}{\linewidth}{@{}l
  >{\hsize=1.5\hsize\centering\arraybackslash}X *{2}{>{\hsize=.7\hsize\centering\arraybackslash}X}
  >{\hsize=1.5\hsize\centering\arraybackslash}X *{2}{>{\hsize=.7\hsize\centering\arraybackslash}X}
  >{\hsize=1.5\hsize\centering\arraybackslash}X >{\hsize=.7\hsize\centering\arraybackslash}X
  >{\hsize=1.0\hsize\columncolor{black!4}\centering\arraybackslash}X@{}}
\toprule
Method & \gd{refA}{\textbf{RefCOCO} val} & \gd{refA}{testA} & \gd{refA}{testB}
 & \gd{refB}{\textbf{RefCOCO+} val} & \gd{refB}{testA} & \gd{refB}{testB}
 & \gd{refC}{\textbf{RefCOCOg} val} & \gd{refC}{test} & \gd{black}{Avg.} \\
\midrule
\ub{Baseline} & \ub{56.2} & \ub{64.4} & \ub{47.5} & \ub{50.1} & \ub{59.4} & \ub{38.9} & \ub{48.8} & \ub{48.4} & \ub{100.0\%} \\
\midrule
\rowcolor{tierbg}\multicolumn{10}{c}{\textbf{Average Tokens = 192}} \\
DART & 32.9 & 36.5 & 30.7 & 28.5 & 33.0 & 26.8 & 29.6 & 29.4 & 60.2\% \\
N\"uwa & \underline{59.0} & \underline{67.3} & \underline{49.0} & \underline{50.5} & \underline{60.8} & \underline{40.2} & \underline{51.7} & \underline{50.1} & \underline{103.6\%} \\
HiPrune & 5.3 & 5.5 & 7.2 & 4.5 & 4.8 & 5.6 & 5.2 & 4.8 & 10.6\% \\
EADP & 7.4 & 7.7 & 9.2 & 6.4 & 7.0 & 7.1 & 6.1 & 5.9 & 14.0\% \\
SID & 5.2 & 5.2 & 6.4 & 4.1 & 4.2 & 5.2 & 4.3 & 4.1 & 9.5\% \\
VLM-Pruner & 46.0 & 54.6 & 39.7 & 39.8 & 48.2 & 32.5 & 38.9 & 39.1 & 81.9\% \\
\rowcolor{oursbg} \textbf{Bifocal (ours)} & \textbf{62.4} & \textbf{71.4} & \textbf{52.9} & \textbf{54.7} & \textbf{66.2} & \textbf{43.1} & \textbf{55.1} & \textbf{54.2} & \textbf{111.0\%} \\
\rowcolor{tierbg}\multicolumn{10}{c}{\textbf{Average Tokens = 128}} \\
DART & 17.0 & 17.5 & 17.1 & 14.8 & 15.7 & 15.1 & 15.4 & 14.6 & 31.0\% \\
N\"uwa & \underline{52.1} & \underline{61.3} & \underline{42.8} & \underline{43.4} & \underline{53.0} & \underline{35.4} & \underline{45.1} & \underline{43.6} & \underline{91.1\%} \\
HiPrune & 3.2 & 3.0 & 4.3 & 2.3 & 2.4 & 3.4 & 2.6 & 2.4 & 5.8\% \\
EADP & 4.2 & 5.0 & 5.8 & 3.1 & 4.2 & 4.6 & 3.5 & 3.4 & 8.3\% \\
SID & 3.0 & 3.0 & 3.9 & 2.6 & 2.8 & 3.5 & 2.5 & 2.4 & 5.9\% \\
VLM-Pruner & 28.3 & 32.4 & 25.5 & 23.9 & 28.3 & 21.5 & 24.7 & 24.8 & 50.8\% \\
\rowcolor{oursbg} \textbf{Bifocal (ours)} & \textbf{57.3} & \textbf{66.4} & \textbf{49.1} & \textbf{49.6} & \textbf{60.3} & \textbf{40.4} & \textbf{49.8} & \textbf{48.7} & \textbf{101.8\%} \\
\rowcolor{tierbg}\multicolumn{10}{c}{\textbf{Average Tokens = 64}} \\
DART & 1.4 & 0.9 & 1.9 & 0.9 & 0.8 & 1.9 & 1.1 & 1.1 & 2.5\% \\
N\"uwa & \underline{32.7} & \underline{39.7} & \underline{24.4} & \underline{26.2} & \underline{33.9} & \underline{20.5} & \underline{25.6} & \underline{24.3} & \underline{54.9\%} \\
HiPrune & 1.3 & 1.3 & 2.1 & 1.1 & 1.1 & 1.6 & 1.3 & 0.9 & 2.6\% \\
EADP & 3.2 & 3.7 & 4.2 & 2.6 & 3.2 & 3.2 & 2.6 & 2.3 & 6.2\% \\
SID & 3.0 & 2.3 & 3.8 & 2.5 & 2.6 & 3.1 & 2.2 & 2.2 & 5.4\% \\
VLM-Pruner & 1.7 & 1.3 & 2.5 & 1.4 & 1.2 & 2.2 & 1.5 & 1.3 & 3.3\% \\
\rowcolor{oursbg} \textbf{Bifocal (ours)} & \textbf{38.6} & \textbf{49.0} & \textbf{30.7} & \textbf{31.9} & \textbf{41.9} & \textbf{24.3} & \textbf{31.0} & \textbf{30.4} & \textbf{66.7\%} \\
\bottomrule
\end{tabularx}
\end{table}

\paragraph{Results on LLaVA-1.5-7B.}
Existing methods are markedly lopsided across tasks. On the general
benchmarks they are hard to tell apart: at the 192-token budget in
Table~\ref{tab:general}, the top five baselines are all crammed between
96.7\% and 98.1\%. Yet they diverge sharply the moment they leave the general
tasks: the OCRBench score of N\"uwa drops from 314 without pruning to 244 at
192 tokens and to only 152 at 64, and on grounding (Table~\ref{tab:grounding})
HiPrune, EADP and SID retain merely 10.6\%, 14.0\% and 9.5\% even at 192
tokens, no longer a mild degradation but a breakdown. Bifocal holds both ends.
It attains the best general average at all three budgets, reaching 98.5\%,
96.3\% and 90.6\%, while on grounding it scores 111.0\% and 101.8\% at 192
and 128 tokens, above the unpruned model, and still retains 66.7\% at 64,
11.8 points above the strongest baseline, N\"uwa. That gap is larger than it
looks, because N\"uwa itself already leads the third-place method, EADP, by
48.7 points at 64 tokens.

\paragraph{Results on Qwen2.5-VL-7B.}
The same pattern persists on a stronger backbone. Qwen2.5-VL is a much better
grounder to begin with, scoring above 80 on all three grounding datasets
without pruning. At the 50\% budget most methods still keep the bulk of it,
but tightening the budget quickly pulls them apart. Under the most aggressive
compression, keeping only 15\% of the tokens, the deduplication-based DART and
VLM-Pruner, which look competitive on general benchmarks at moderate budgets,
fall to 6.9\% and 9.5\% of the unpruned grounding accuracy, and VisionZip,
EADP and HiPrune all hover around 40\%. As shown in Table~\ref{tab:qwen},
Bifocal ranks first in both general and grounding averages at all three token
budgets, and it leads on every grounding split
(\apx{Appendix~\ref{app:qwen_grounding}}). Its lead also widens exactly where
pruning hurts most. At 15\%, Bifocal still keeps about two thirds of the
unpruned grounding accuracy, while no baseline retains even 60\%.

\begin{table}[ht]
\centering
\caption{\textbf{Results on Qwen2.5-VL-7B.} Budgets are fractions of the visual
tokens. Gen.\ and REC are the performance relative to the baseline on the
general and grounding benchmarks.}
\label{tab:qwen}
\scriptsize
\setlength{\tabcolsep}{2.2pt}\renewcommand{\arraystretch}{0.92}
\setlength{\aboverulesep}{0pt}\setlength{\belowrulesep}{0pt}
\begin{adjustbox}{max width=\linewidth}
\begin{tabular}{@{}l *{11}{c} *{2}{>{\columncolor{black!4}}c}@{}}
\toprule
\multirow{2}{*}{Method} & \cat{2}{catGen}{General} & \cat{2}{catRea}{Reason.} & \cat{3}{catOcr}{OCR} & \cat{1}{catHal}{Hall.} & \cat{3}{refA}{Grounding} & \cat{2}{black}{Avg.} \\
\cmidrule(lr){2-3}\cmidrule(lr){4-5}\cmidrule(lr){6-8}\cmidrule(lr){9-9}\cmidrule(lr){10-12}\cmidrule(lr){13-14}
 & GQA & MMB & SQA & AI2D & VQA\textsuperscript{T} & OCR-B & DocVQA & POPE & RefCOCO & RefCOCO+ & RefCOCOg & Gen. & REC \\
\midrule
\ub{Baseline} & \ub{60.9} & \ub{84.0} & \ub{88.2} & \ub{82.3} & \ub{82.8} & \ub{842} & \ub{94.7} & \ub{86.4} & \ub{88.7} & \ub{82.6} & \ub{84.1} & \ub{100.0\%} & \ub{100.0\%} \\
\midrule
\rowcolor{tierbg}\multicolumn{14}{c}{\textbf{Retain 50\% Tokens}} \\
VisionZip & 57.4 & 78.7 & 79.6 & 73.5 & 74.8 & 734 & 89.5 & \underline{86.5} & 77.8 & 74.9 & 77.5 & 91.5\% & 89.6\% \\
DART & 59.3 & 81.5 & 86.2 & 78.0 & 79.5 & 721 & 77.5 & 85.6 & 76.1 & 69.8 & 72.5 & 93.1\% & 85.5\% \\
N\"uwa & \textbf{60.1} & \textbf{83.7} & \textbf{87.6} & 81.6 & \underline{81.4} & 813 & \underline{93.2} & 86.0 & \underline{86.7} & \underline{79.6} & \underline{81.9} & \underline{97.6\%} & \underline{97.2\%} \\
HiPrune & 57.3 & 79.2 & 79.3 & 74.0 & 74.8 & 731 & 89.3 & \textbf{86.6} & 77.5 & 74.7 & 77.7 & 91.7\% & 89.4\% \\
EADP & 59.5 & 82.6 & 85.5 & 80.0 & 77.3 & 741 & 81.0 & 84.9 & 82.1 & 74.5 & 76.7 & 93.4\% & 91.5\% \\
VLM-Pruner & \underline{59.9} & \underline{83.0} & \underline{87.3} & \textbf{82.1} & 81.0 & \textbf{820} & 91.3 & 85.8 & 83.7 & 77.2 & 79.2 & 97.4\% & 94.0\% \\
\rowcolor{oursbg} \textbf{Bifocal (ours)} & \textbf{60.1} & 82.6 & \underline{87.3} & \underline{82.0} & \textbf{82.8} & \underline{814} & \textbf{94.0} & 86.3 & \textbf{87.5} & \textbf{80.7} & \textbf{82.8} & \textbf{98.4\%} & \textbf{98.3\%} \\
\rowcolor{tierbg}\multicolumn{14}{c}{\textbf{Retain 25\% Tokens}} \\
VisionZip & 55.1 & 77.0 & 78.3 & 70.0 & 68.6 & 601 & 77.5 & 84.0 & 61.2 & 58.5 & 59.4 & 84.8\% & 69.9\% \\
DART & 55.7 & 77.3 & 82.8 & 70.1 & 66.4 & 528 & 48.7 & 80.8 & 26.6 & 24.1 & 25.9 & 80.3\% & 29.9\% \\
N\"uwa & 58.5 & \textbf{82.4} & \underline{86.7} & \underline{81.0} & \underline{77.9} & \underline{680} & \underline{87.8} & \underline{85.7} & \underline{77.0} & \underline{68.2} & \underline{72.7} & \underline{93.4\%} & \underline{85.4\%} \\
HiPrune & 54.9 & 77.7 & 77.7 & 70.2 & 68.8 & 604 & 78.4 & 84.5 & 59.3 & 56.3 & 58.7 & 85.1\% & 67.9\% \\
EADP & 58.6 & 79.5 & 83.7 & 76.2 & 70.8 & 641 & 66.7 & 83.3 & 61.0 & 55.7 & 56.3 & 86.8\% & 67.9\% \\
VLM-Pruner & \underline{58.9} & 80.7 & 85.7 & 78.5 & 76.3 & 678 & 76.8 & 84.9 & 44.1 & 40.5 & 40.1 & 90.3\% & 49.1\% \\
\rowcolor{oursbg} \textbf{Bifocal (ours)} & \textbf{59.5} & \underline{82.0} & \textbf{87.2} & \textbf{81.3} & \textbf{80.3} & \textbf{711} & \textbf{90.3} & \textbf{86.2} & \textbf{81.2} & \textbf{72.0} & \textbf{77.0} & \textbf{94.9\%} & \textbf{90.2\%} \\
\rowcolor{tierbg}\multicolumn{14}{c}{\textbf{Retain 15\% Tokens}} \\
VisionZip & 53.3 & 75.1 & 76.7 & 66.1 & 60.8 & 463 & 60.3 & 82.1 & 38.9 & 36.3 & 36.6 & 77.0\% & 43.8\% \\
DART & 50.8 & 71.4 & 80.0 & 66.5 & 49.9 & 383 & 30.1 & 70.1 & 6.3 & 5.5 & 5.8 & 68.0\% & 6.9\% \\
N\"uwa & \underline{56.2} & \textbf{80.5} & \underline{85.2} & \underline{77.6} & \underline{72.2} & 540 & \underline{75.5} & \underline{84.0} & \underline{53.7} & \underline{44.3} & \underline{49.4} & \underline{86.7\%} & \underline{58.0\%} \\
HiPrune & 52.5 & 75.0 & 76.6 & 66.9 & 62.8 & 457 & 64.0 & 82.3 & 35.7 & 33.0 & 34.4 & 77.7\% & 40.3\% \\
EADP & \textbf{57.7} & 78.0 & 82.3 & 72.7 & 65.5 & \underline{564} & 56.8 & 80.9 & 36.6 & 33.7 & 34.1 & 81.4\% & 40.9\% \\
VLM-Pruner & 55.2 & 76.0 & 83.5 & 71.1 & 66.5 & 487 & 53.2 & 78.5 & 8.5 & 8.0 & 7.8 & 77.8\% & 9.5\% \\
\rowcolor{oursbg} \textbf{Bifocal (ours)} & \textbf{57.7} & \underline{79.5} & \textbf{85.8} & \textbf{77.9} & \textbf{76.3} & \textbf{578} & \textbf{81.1} & \textbf{85.3} & \textbf{62.7} & \textbf{52.4} & \textbf{57.5} & \textbf{89.0\%} & \textbf{67.9\%} \\
\bottomrule
\end{tabular}
\end{adjustbox}
\end{table}



\subsection{Ablation Study}
\label{sec:ablation}

Bifocal selects and allocates three kinds of tokens from different angles, so
we verify that each of them is indeed useful. Taking LLaVA-1.5-7B with an
average of $128$ tokens per LLM layer as an example, we vary how the $224$
Stage-1 tokens are split among $C$, $I$ and $Q$ (Table~\ref{tab:routes}).
$C$ clearly pays off on grounding: adding it lifts REC from 83.6\% to 89.0\%
with $I$ alone, and from 97.0\% to 101.8\% with $Q$ alone. Like $I$, $C$ is
query-free, so the general average changes little. As expected, removing $I$
lowers POPE from 84.7 to 83.2, and removing $Q$ drops REC from 102.7\% to
89.0\%. We also note that the full configuration is not the best on every
benchmark, which is a matter of balance. As analysed in
Section~\ref{sec:sources}, keeping all three kinds of tokens yields the best
overall ability, making it the only allocation that leads on both averages,
with 95.4\% on general and 102.7\% on grounding.

\begin{table}[ht]
\centering
\caption{\textbf{Evidence ablation}.
$C$, $I$ and $Q$ are the numbers of coarse-context, image-intrinsic and
query-focused tokens, all rows apply the same Stage-2
pruning. }
\label{tab:routes}
\scriptsize
\providecommand{\abz}{\textcolor{black!30}{--}}
\setlength{\tabcolsep}{2pt}\renewcommand{\arraystretch}{1.0}
\setlength{\aboverulesep}{0pt}\setlength{\belowrulesep}{0pt}
\begin{tabularx}{\linewidth}{@{}*{3}{>{\centering\arraybackslash}p{0.62cm}}
  *{10}{>{\centering\arraybackslash}X}
  *{2}{>{\columncolor{black!4}\centering\arraybackslash}X}@{}}
\toprule
\multicolumn{3}{c}{\textbf{Token allocation}} & \cat{2}{catGen}{General} & \cat{2}{catRea}{Reasoning} & \cat{3}{catOcr}{OCR} & \cat{1}{catHal}{Hall.} & \cat{2}{refB}{RefCOCO+} & \cat{2}{black}{Avg.} \\
\cmidrule(lr){1-3}\cmidrule(lr){4-5}\cmidrule(lr){6-7}\cmidrule(lr){8-10}\cmidrule(lr){11-11}\cmidrule(lr){12-13}\cmidrule(lr){14-15}
$C$ & $I$ & $Q$ & GQA & MMB & SQA & AI2D & OCR-B & DocVQA & ChartQA & POPE & testA & testB & Gen. & REC \\
\midrule
\multicolumn{3}{c}{\ub{Baseline(576)}} & \ub{61.9} & \ub{64.1} & \ub{69.4} & \ub{55.2} & \ub{314} & \ub{21.5} & \ub{18.2} & \ub{85.9} & \ub{59.4} & \ub{38.9} & \ub{100.0\%} & \ub{100.0\%} \\
\midrule
\abz & 224 & \abz & \textbf{60.6} & 63.8 & 68.2 & \textbf{54.8} & 200 & 20.9 & 17.3 & 84.3 & 49.7 & 32.5 & 93.6\% & 83.6\% \\
\abz & 112 & 112 & 59.8 & 63.9 & \textbf{68.3} & 53.1 & 213 & \textbf{21.2} & \textbf{17.4} & 84.2 & 55.0 & 38.3 & 93.9\% & 95.5\% \\
\abz & \abz & 224 & 59.6 & 63.7 & 68.0 & 51.9 & 242 & 20.8 & 16.8 & 83.7 & 56.5 & 38.5 & 93.9\% & 97.0\% \\
\midrule
64 & 160 & \abz & 60.3 & 63.7 & 68.2 & 54.6 & 209 & 19.4 & 17.3 & 83.7 & 53.1 & 34.4 & 92.9\% & 89.0\% \\
\rowcolor{oursbg} \textbf{64} & \textbf{80} & \textbf{80} & 60.5 & \textbf{64.1} & 68.2 & 54.2 & \textbf{262} & 20.4 & 16.8 & \textbf{84.7} & \textbf{60.3} & 40.4 & \textbf{95.4\%} & \textbf{102.7\%} \\
64 & \abz & 160 & 60.0 & 63.1 & 67.7 & 53.5 & 226 & 19.3 & 16.3 & 83.2 & 58.6 & \textbf{40.8} & 92.2\% & 101.8\% \\
\bottomrule
\end{tabularx}
\end{table}

\subsection{Visualisation}
\label{sec:visual-study}

\begin{figure}[ht]
  \centering
  \includegraphics[width=\linewidth]{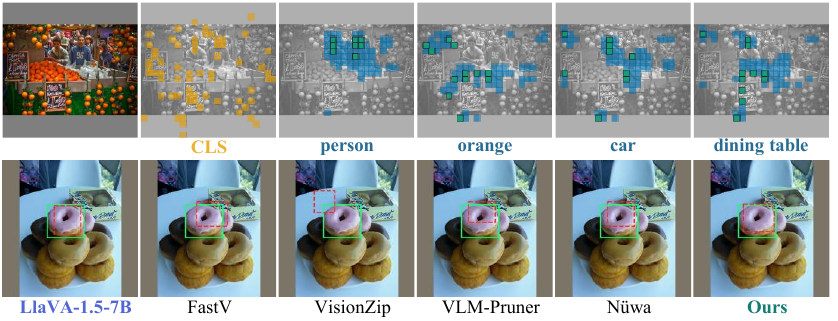}
  \caption{\textit{Top:} the fine evidence Bifocal
  selects for one image under different queries, \mkfine~image-intrinsic
  evidence and \sw{routeTxt}~query-focused evidence. \textit{Bottom:} grounding
  results of LLaVA-1.5-7B and pruning methods under the same budget, with
  \mkgt~the ground truth and \mkpred~the prediction.}
  \label{fig:visual}
\end{figure}

We further examine Bifocal qualitatively in Figure~\ref{fig:visual}, from two
angles. The top row visualises the two kinds of fine evidence Bifocal keeps for one image under different queries, with more cases in
\apx{Appendix~\ref{app:visual}}. The image-intrinsic evidence spreads over the
whole image and stays the same whatever the question, while the query-focused
evidence moves with the query. For \textit{person} and \textit{orange}, two
objects present in the image, the query-focused tokens land on the region each
query names. The two kinds of fine evidence thus split the work as intended,
one covering the image and the other following the question. The bottom row
turns to fine-grained ability. We visualise both grounding and OCR, and show
one grounding case here, with the rest in
\apx{Figures~\ref{fig:visual-qwen} and~\ref{fig:visual-llava}}. The image
holds several donuts that easily mislead the model, yet Bifocal's prediction is
clearly better than those of the other methods in both centre and IoU.

\FloatBarrier
\section{Conclusion}
\label{sec:conclusion}

This paper approaches visual token pruning from a novel perspective and
proposes a mixed-granularity pruning framework before the LLM, composed of
query-focused evidence, image-intrinsic evidence and a coarse context, which
preserves the abilities of the model more comprehensively. Building on this
framework, we optimise the selection of each kind of token. We combine the word
and image saliency maps to better preserve query-focused evidence through word
weighting and word conjunction, then select the optimal token subset by
representational coverage, and finally add a coarse context pooled from the
remaining patches. Together with Stage~2, which further removes redundant
tokens inside the LLM, Bifocal achieves the best overall performance across
multiple models and budgets.


\clearpage
\bibliographystyle{plainnat}
\bibliography{references}

\clearpage
\appendix

\section{The Contribution of Each Evidence Route} \label{app:single}

The main text (\S\ref{sec:sources}) concludes that the two evidence routes fail in different ways and cannot substitute for each other. In this appendix we discuss the contribution of each route in turn.

\subsection{Which route matches the intuition of the dense model?}
\label{app:align}
\begin{figure}[htbp]
  \centering
  \includegraphics[width=\linewidth]{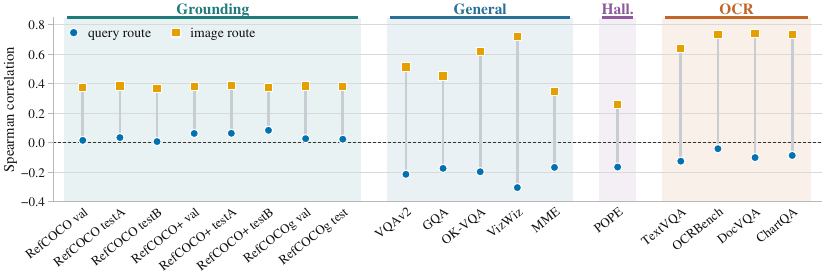}
  \caption{\textbf{Alignment of the image and text routes with the baseline's
  attention.} \legmk{routeTxt}{\bullet}~text route. \legmk{routeImg}{\scriptstyle\blacksquare}~image route. Error bars are $95\%$ confidence intervals.}
  \label{fig:align}
\end{figure}

We first look at which evidence a dense model relies on when it solves a task
normally. For each sample, let $A_j$ be the attention that the dense model pays
to visual token $j$ while generating its answer, and $s^{r}_j$ the score that
route $r$ assigns to it. Since pruning only depends on the ranking of the
scores, we measure their agreement by the Spearman correlation over the $576$
visual tokens:
\begin{equation}
\rho^{r}=\mathrm{corr}\big(\mathrm{rank}(A),\ \mathrm{rank}(s^{r})\big),
\qquad r\in\{\text{text},\ \text{image}\}.
\label{eq:align}
\end{equation}
Figure~\ref{fig:align} shows a clear trend. The image route is positively
correlated with the attention of the dense model on all $18$ tasks, and is
higher than the text route on $92\%$ to $100\%$ of the samples in every task.
The strength of this agreement, however, varies with the task: it is around
$0.38$ on grounding and reaches $0.74$ on OCR and document tasks. The text
route stays much weaker, slightly positive only on grounding and negative on
the other tasks. The image thus lays the foundation of what the dense model
reads, and different tasks attend to it to different degrees. Although the
text route agrees less with the dense model, it is still necessary: as the
next subsection shows, on grounding it decides whether the right tokens are
kept at all.

\subsection{The text route: a necessary condition for grounding}
\begin{figure}[htbp]
  \centering
  \includegraphics[width=\linewidth]{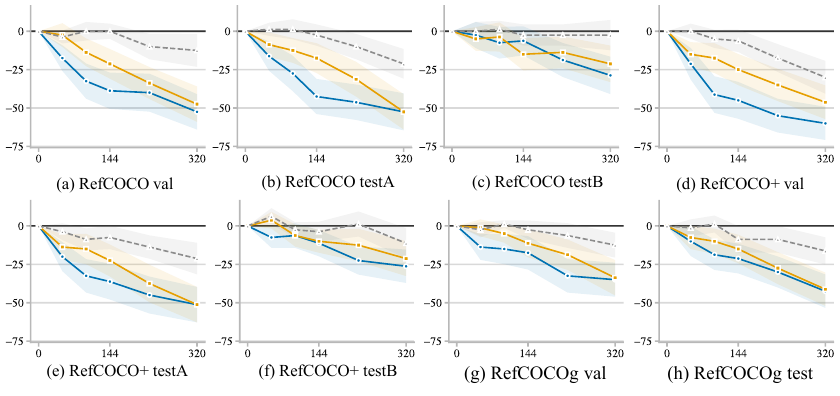}
  \caption{\textbf{Deletion test on the eight REC splits}. The horizontal axis is the number of tokens removed from the
  unpruned $576$. The vertical axis is the change in \textsc{Acc}@$0.5$
  relative to keeping all $576$, in percentage points. Each panel shows
  three curves:
  \capleg{routeTxt}{\bullet}{text route},
  \capleg{routeImg}{\scriptstyle\blacksquare}{image route}, and
  \caplegd{routeRand}{\scriptstyle\blacktriangle}{random}.}
  \label{fig:drop}
\end{figure}

For the text route, we use a deletion test to ask the most direct question: can we do without it? Starting from the unpruned sequence of $576$ visual tokens, we remove equal numbers of tokens selected by the text route, the image route, or at random, while progressively increasing the number removed. Figure~\ref{fig:drop} shows a consistent trend across eight REC splits: at low removal rates, random deletion has only a limited effect, whereas removing text-route tokens causes \textsc{Acc}@$0.5$ to drop sharply, usually faster than removing image-route tokens. The text route may not carry the majority of the evidence, but it provides query-specific evidence that grounding cannot do without---if the model does not know what the question refers to, how can it draw the box accurately? As the removal rate approaches one half, the curves of the two evidence routes gradually converge. At this point, large-scale information loss begins to dominate, and which route the removed tokens come from is no longer the main issue. This further supports our conclusion in the main text: the two evidence routes cannot substitute for each other and should each be allocated their own share of the fixed budget.

\subsection{The image route: a wider view}
However well the text route is optimised, what it keeps is still word-level
evidence: before the LLM there is no real understanding of the question, so
the route can only keep the patches that look like the words it was given.
This fails in two ways: the object the question refers to is not in the image
at all, or the answer sits on something the question never names (or the
question has no target to begin with). In both cases, selecting tokens purely
by word similarity means that, under a reduced budget, the model is shown the
wrong evidence, and hallucination follows. This is exactly what the image
route is for: it does not read the question, and its wider view brings a more
diverse sample of the scene into the budget.

\begin{figure}[ht]
  \centering
  \includegraphics{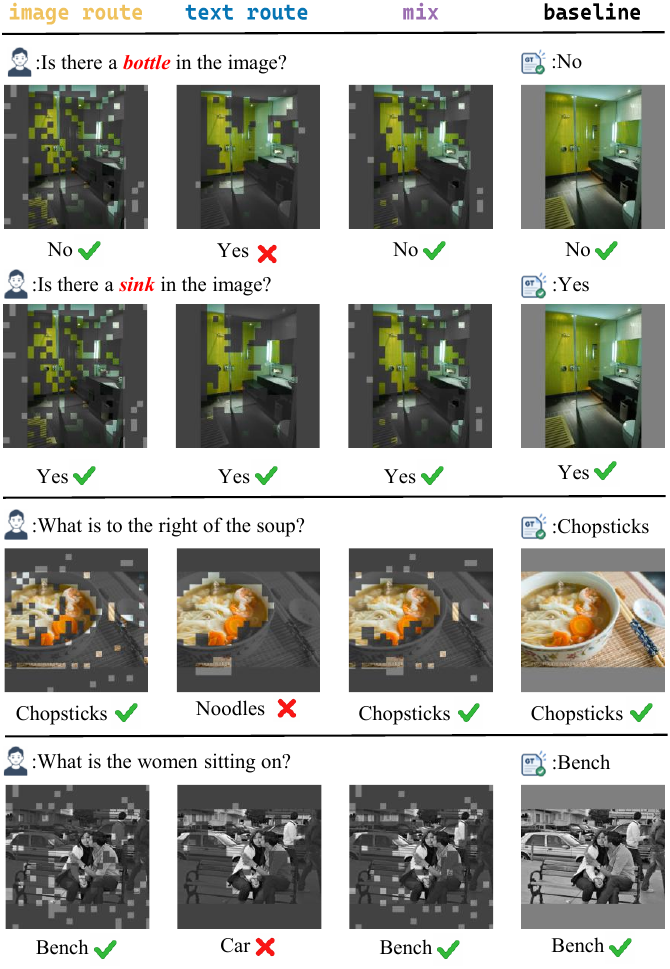}
  \caption{\textbf{Answers under single-route inputs.} Each row is one
  question, with the ground truth shown by the icon on the right. The four
  columns are four visual token inputs:
  \textcolor{routeImg}{image route} (CLS top-$144$),
  \textcolor{routeTxt}{text route} (scoring-chain top-$144$),
  \textcolor{routeBoth}{mix} (half from each route), and baseline (all $576$ tokens).
  Within each panel, kept tokens retain their colour and discarded ones are
  dimmed.}
  \label{fig:case-cls}
\end{figure}

Figure~\ref{fig:case-cls} shows three cases. In the first, the same image is
asked two different questions: the tokens selected by the text route change
with the question, while the image route does not read the query and its two
rows are identical patch for patch. When asked about a bottle that is not in
the image, only the text route answers ``yes'', while the image route and the
half-and-half mix both answer correctly. When asked about the sink that is
present, all four inputs answer correctly---so the text route does not say
``yes'' to everything, it says ``yes'' on fragments that merely look like the
queried object. The remaining two cases show the other failure: what the
question names is not where the answer lies, the text route again answers
wrong, and the image route is needed to bring that wider view back.

\section{Plugging Existing Pruning Methods into the Framework}
\label{app:plug}

To verify whether the mixed-granularity framework is effective in itself, we
plug existing pruning methods into it (Table~\ref{tab:plug}). Each method
replaces the selection rule of only one kind of fine evidence, while the
budgets of all three kinds of tokens stay fixed, and is compared with the same
method used on its own under the same total budget. For image-intrinsic
evidence, HiPrune and VisionZip replace our image-evidence selection on
LLaVA-1.5-7B. Grounding on LLaVA-1.5-7B hardly separates different
query-focused methods, so for query-focused evidence EADP replaces our
query-evidence selection on Qwen2.5-VL-7B. With the mixed-granularity
framework, every method improves substantially, most clearly on grounding:
the per-dataset gains range from $9.7$ to $20.6$ points, and Avg.\ rises by
$6.8$ to $9.7$ points.

\providecommand{\gain}[1]{\,{\tiny\textcolor{catRea}{#1}}}
\begin{table}[ht]
\centering
\caption{\textbf{Existing methods with and without our framework.} HiPrune
and VisionZip select the image-intrinsic evidence within the framework on
LLaVA-1.5-7B, and EADP selects the query-focused evidence on Qwen2.5-VL-7B.
Each grounding column averages \textsc{Acc}@$0.5$ over the val and test splits
of that dataset, and Avg.\ averages all metrics relative to the unpruned model.}
\label{tab:plug}
\scriptsize
\setlength{\tabcolsep}{3pt}\renewcommand{\arraystretch}{1.1}
\setlength{\aboverulesep}{0pt}\setlength{\belowrulesep}{0pt}
\begin{adjustbox}{max width=\linewidth}
\begin{tabular}{l *{10}{c} >{\columncolor{black!4}}c}
\toprule
\multirow{2}{*}{Method} & \cat{2}{catGen}{General} & \cat{2}{catRea}{Reasoning} & \cat{2}{catOcr}{OCR} & \cat{1}{catHal}{Hall.} & \cat{3}{refA}{Grounding} & \cat{1}{black}{Avg.} \\
\cmidrule(lr){2-3}\cmidrule(lr){4-5}\cmidrule(lr){6-7}\cmidrule(lr){8-8}\cmidrule(lr){9-11}\cmidrule(lr){12-12}
 & GQA & MMB & SQA-I & AI2D & VQA\textsuperscript{T} & OCR-B & POPE & RefCOCO & RefCOCO+ & RefCOCOg & \\
\midrule
\rowcolor{tierbg}\multicolumn{12}{c}{\textbf{LLaVA-1.5-7B}, image-intrinsic evidence} \\
HiPrune & 58.2 & 64.0 & 68.3 & 53.0 & 44.2 & 290 & 81.2 & 16.3 & 13.9 & 15.2 & 75.9\% \\
\rowcolor{oursbg}\quad + framework & 59.3 & 63.3 & 68.2 & 52.9 & 43.8 & 288 & 80.5 & \textbf{36.9}\gain{+20.6} & \textbf{30.7}\gain{+16.8} & \textbf{28.6}\gain{+13.4} & \textbf{85.6\%}\gain{+9.7} \\
VisionZip & 58.4 & 63.4 & 68.2 & 52.3 & 44.6 & 301 & 81.1 & 26.4 & 22.6 & 22.8 & 81.3\% \\
\rowcolor{oursbg}\quad + framework & 59.2 & 63.6 & 68.1 & 52.7 & 44.5 & 297 & 80.4 & \textbf{41.5}\gain{+15.1} & \textbf{34.8}\gain{+12.2} & \textbf{32.5}\gain{+9.7} & \textbf{88.5\%}\gain{+7.2} \\
\midrule
\rowcolor{tierbg}\multicolumn{12}{c}{\textbf{Qwen2.5-VL-7B}, query-focused evidence} \\
EADP & 58.6 & 79.5 & 83.7 & 76.2 & 70.8 & 641 & 83.3 & 61.0 & 55.7 & 56.3 & 84.0\% \\
\rowcolor{oursbg}\quad + framework & 58.8 & 81.7 & 85.8 & 80.2 & 78.2 & 703 & 84.7 & \textbf{73.4}\gain{+12.4} & \textbf{66.6}\gain{+10.9} & \textbf{67.4}\gain{+11.1} & \textbf{90.8\%}\gain{+6.8} \\
\bottomrule
\end{tabular}
\end{adjustbox}
\end{table}

\section{Per-Cell Noise in the Cosine Map and What Smoothing Removes}
\label{app:smooth}

The starting point is simple. Pruning keeps the top-$m$ cells by score, and
$576$ cells compete for that small handful of seats at the upper tail. If the
score of each cell has picked up the perturbation that comes with computing a
cosine, the quality of the top-$k$ suffers. This section is therefore about
three things: that the noise exists, that a simple averaging operation can
filter it out, and that this operation is safe, since the low-pass filter does
not remove the information that matters.

Which cells of a cosine map are content and which are jitter can be told
apart by a property that requires no labels: the two spread over different
ranges. Content is contiguous---an object always occupies several cells, so
the values of neighbouring cells are correlated---whereas estimation jitter
is independent from cell to cell and is already unrelated one cell away. The
vague question of whether there is noise, and how much, thus becomes a
measurable one: how much of the variance in the map lives only at the scale
of a single cell.

We therefore look for evidence in two directions. One is spatial: how much
relation a token still has with the cells around it. The other is spectral:
among the light-and-dark fluctuations on the map, how many are large patches
that vary slowly, and how many vary from one cell to the next? Objects are
contiguous, so the fluctuations they produce can only be the former, that is,
concentrated at low frequency.

\begin{figure}[htbp]
  \centering
  \includegraphics[width=\linewidth]{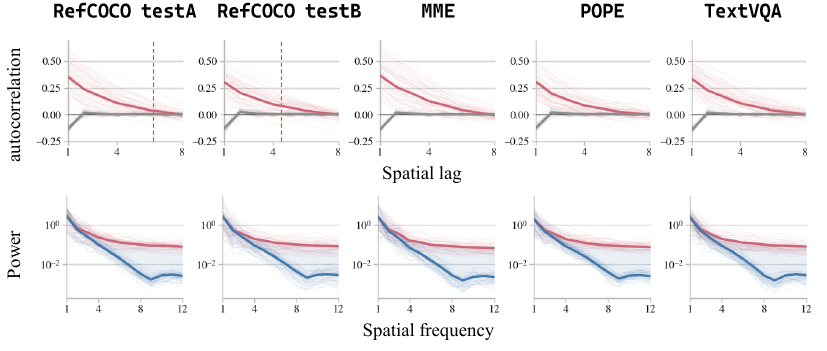}
  \caption{Top row: autocorrelation against spatial lag, in cells. Bottom
  row: radially averaged power spectrum against spatial frequency, on a
  logarithmic vertical axis. Curves are \sw{mapRaw}~the raw map $s_w$,
  \sw{mapSm}~the map after the $3{\times}3$ mean filter $\tilde s_w$, and
  \sw{mapRes}~the removed component $s_w-\tilde s_w$. Thin lines are
  individual samples and thick lines their mean. The dashed vertical line
  marks the median half-width of the ground-truth boxes.}
  \label{fig:smooth-scale}
\end{figure}

Figure~\ref{fig:smooth-scale} shows both. The first is the spatial one, which
plots autocorrelation against distance. The horizontal axis is the distance
$d$ between two cells (in cells), and the vertical axis is the correlation
coefficient of the values of two cells that far apart, averaged over samples,
\[
  \rho(d)=\frac{\operatorname{Cov}\big(s(i),\,s(j)\big)}{\operatorname{Var}(s)},
  \qquad \lVert i-j\rVert = d .
\]
The red curve is $\rho(d)$ of the raw per-word map. The correlation of a cell
with itself is always $1$. It decreases with distance, and once the gap
between two cells exceeds the scale of the content itself it falls to zero.
What we actually read is this: the red curve has already dropped to about a
half at $d=1$, and only then decays gently, reaching zero at around six
cells, which matches the median half-width of the ground-truth boxes (dashed
line). The initial drop and the subsequent decay say two different things.
Writing the per-word map as content plus noise, $s=c+\varepsilon$ with
$\varepsilon$ independent from cell to cell, $\varepsilon$ contributes nothing
to the covariance at $d\ge 1$, so
\[
  \rho(d)=(1-\alpha)\,\rho_c(d)\quad (d\ge 1),
  \qquad \alpha=\frac{\operatorname{Var}(\varepsilon)}{\operatorname{Var}(s)} .
\]
That is, the drop from $1$ to $\rho(1)$ is exactly the share of variance
$\alpha$ held by the cell-independent component, while the distance over
which the gentle part extends is the correlation length of the content. The
grey curve is the autocorrelation of the removed component: one cell out it
already sits at zero, or even slightly below---it is unrelated even to its own
neighbours and cannot carry content.

The second figure is another way of looking at the same thing.
Autocorrelation splits the variance by distance, the power spectrum splits
the same variance by scale, and the two are strict Fourier pairs
(Wiener--Khinchin theorem). Concretely, a two-dimensional discrete Fourier
transform of the $24\times24$ per-word map gives coefficients $F(u)$, and the
energy at a given frequency is $|F(u)|^2$, which we then average radially over
the modulus of the frequency:
\[
  P(f)=\big\langle\,|F(u)|^2\,\big\rangle_{\lVert u\rVert = f}.
\]
Summing the energy over all frequencies gives exactly the total variance of
the map (Parseval's identity), so asking which frequencies the energy falls
on is asking how coarse the fluctuations that make up this variance are: the
horizontal axis is spatial frequency, where low frequency corresponds to
large, slow variation and high frequency to jitter that changes cell by cell.
The vertical axis is logarithmic because the magnitudes are too far apart---
the energy of the content at low frequency is one to two orders above the
noise plateau, and on a linear axis the plateau would be squashed against
zero, hiding the fact that it is flat.

The red curve is markedly high at low frequency, which says the map does
carry contiguous content, while at high frequency it flattens into a horizontal
plateau. Flatness means equal energy at every frequency, which is the
signature of cell-independent noise, and the height of the plateau is the
power of that noise. The blue curve is the spectrum after smoothing: at low
frequency the two curves almost coincide, while at high frequency it is an
order of magnitude lower---what has been removed is exactly that plateau, and
the low frequency where the content lives is barely touched.

The other two curves in the figure confirm this. The grey curve is the
removed part, and one cell out it is unrelated to its neighbours, which says
that what we dropped is not content. The blue curve is the spectrum after
smoothing, pushed down only at high frequency and almost coinciding with the
red curve at low frequency, which says the content at low frequency is still
there as it was. The $3\times3$ filter is therefore not smearing detail away
but removing the jitter that should not win a seat: the part it removes
carries no content, and the step is safe.

\section{Norm Bias in the CLS Attention}
\label{app:norm}

The image saliency map $S$ of Section~\ref{sec:visual} is the residual of the
smoothed CLS attention after a linear regression on the patch norm
(Eq.~\eqref{eq:S}). This appendix explains why this correction is needed. We
first derive why the CLS attention follows the norm and verify each step of the
derivation empirically (Appendix~\ref{app:norm-why}), and then show how
removing the norm-explained component changes the patches ranked at the top
(Appendix~\ref{app:desink}).

\subsection{Why the CLS Attention Follows the Norm}
\label{app:norm-why}

LayerNorm rescales every patch feature to the same length before the attention,
so it is not obvious why the norm should affect the CLS attention at all. The
CLS logit for patch $j$ is the inner product between the CLS query $q$ and the
key, and the key is a linear transform of the normalised feature $z_j$.
LayerNorm fixes the length of $z_j$ but not its composition. A large norm comes
from a few channels with massive
activations~\citep{sun2024massive,darcet2024vision}, which point along a shared
direction $v$ that is stable across images. We call it the \emph{high-norm
direction}. We therefore split $z_j$ into its component along $v$ and the rest:
\begin{equation}
z_j=\|z\|\big(\rho_j\,v+\sqrt{1-\rho_j^2}\,u_j\big),
\label{eq:split}
\end{equation}
where $\rho_j=\cos(z_j,v)$, which we call the \emph{norm alignment} of patch
$j$, and $u_j\perp v$ is the direction of the remainder. The larger the norm,
the more these channels weigh after normalisation, and the higher the norm
alignment: LayerNorm removes the norm, but the norm survives as alignment.

Since both the key projection and the inner product are linear, substituting
this split separates the logit in each attention head into two terms:
\begin{equation}
\mathrm{logit}_j=\frac{q^\top W_k z_j}{\sqrt d}
=\rho_j\,\frac{\|z\|\,q^\top W_k v}{\sqrt d}
+\sqrt{1-\rho_j^2}\,\frac{\|z\|\,q^\top W_k u_j}{\sqrt d}
=G\,\rho_j+g_j.
\label{eq:logit}
\end{equation}
The first term is where the norm enters. $G$ is shared by all patches, so
$G\rho_j$ is a bonus that each patch receives in proportion to its norm
alignment, regardless of what it contains, and only $g_j$ reflects the content. The
softmax then turns this sum into a product:
\begin{equation}
a_j=\frac{e^{\mathrm{logit}_j}}{\sum_k e^{\mathrm{logit}_k}}=n_j\,c_j,
\qquad n_j=e^{G\rho_j},\quad c_j=\frac{e^{g_j}}{Z},
\label{eq:product}
\end{equation}
where $Z=\sum_k e^{G\rho_k+g_k}$ is the softmax normaliser, $n_j$ is a gain set
by the norm alignment, and $c_j$ is the content score.
This is why the CLS attention follows the norm: a patch of large norm is
multiplied by a large gain, whatever it contains. The derivation only needs $G$
to exist, and its sign and size differ from head to head.

To verify this account, we feed all images of GQA, TextVQA, RefCOCO testA and
RefCOCO testB to LLaVA-1.5-7B and measure, on the ViT layer used for scoring,
three things: how strongly the CLS attention is biased toward high-norm
patches, whether the bias arises through the mechanism derived above, and what
the fit of Eq.~\eqref{eq:S} does to the map. Table~\ref{tab:normcheck} reports
the medians over images. Throughout, $\mathcal H$ denotes the eight highest-norm
patches of an image, $1.4\%$ of the grid.

\paragraph{Bias.}
$\pi$ is the share of the CLS attention received by $\mathcal H$, and $\kappa$ is
the ratio of its mean attention to that of an ordinary patch. $r$ and
$r_{\setminus\mathcal H}$ are the Pearson correlations between the attention map
$\tilde a$ and the norm map $\nu$, over all patches and with $\mathcal H$
excluded. The bias is large: $\mathcal H$ receives $17\%$--$22\%$ of the
attention, $14$ to $19$ times an ordinary patch. It is also not confined to
$\mathcal H$, since the correlation drops only slightly once $\mathcal H$ is
excluded. We therefore remove the whole norm-explained component rather than a
few patches.

\paragraph{Mechanism.}
$\gamma$ is the mean cosine between the high-norm directions $v$ of different
images, and $R^2_\rho$ and $R^2_{\|h\|}$ are the shares of the logit variance
explained by a linear fit on the norm alignment $\rho_j$ and on the patch norm
$\|h_j\|$. With $\gamma=0.96$ on every split, $v$ is a property of the model
rather than of the image. $R^2_\rho$ is about four times $R^2_{\|h\|}$, so the
logit follows the norm alignment rather than the norm itself, as the derivation
predicts once LayerNorm has removed the length.

\paragraph{Correction.}
Taking the top $5\%$ of patches by $\nu$ as high-norm, $\eta_{\tilde a}$ and
$\eta_S$ are their share of the top $10\%$ of $\tilde a$ and of $S$, where chance
is $10\%$. $r_s$ is the Spearman correlation between $\tilde a$ and $S$. The fit
of Eq.~\eqref{eq:S} lowers the high-norm share from $64\%$--$75\%$ to
$46\%$--$50\%$, while $r_s\ge 0.87$ shows that the ranking of the other patches
stays largely intact. The share remains above chance because the correction is
first-order: sinks that rise far above the fitted line stay on top.

\begin{table}[!ht]
\centering
\caption{\textbf{Empirical check of the norm bias} on LLaVA-1.5-7B, median over
all images of each split. Symbols are defined in the text.}
\label{tab:normcheck}
\small
\setlength{\tabcolsep}{3pt}
\begin{tabularx}{\linewidth}{@{}l *{10}{>{\centering\arraybackslash}X}@{}}
\toprule
 & \multicolumn{4}{c}{Bias} & \multicolumn{3}{c}{Mechanism} & \multicolumn{3}{c}{Correction} \\
\cmidrule(lr){2-5}\cmidrule(lr){6-8}\cmidrule(lr){9-11}
Split & $\pi$ & $\kappa$ & $r$ & $r_{\setminus\mathcal H}$ & $\gamma$ & $R^2_\rho$ & $R^2_{\|h\|}$ & $\eta_{\tilde a}$ & $\eta_S$ & $r_s$ \\
\midrule
GQA           & 22\% & 19$\times$ & 0.53 & 0.49 & 0.96 & 0.67 & 0.17 & 75\% & 46\% & 0.87 \\
TextVQA       & 17\% & 14$\times$ & 0.30 & 0.25 & 0.96 & 0.59 & 0.16 & 64\% & 46\% & 0.92 \\
RefCOCO testA & 21\% & 18$\times$ & 0.53 & 0.48 & 0.96 & 0.67 & 0.18 & 68\% & 46\% & 0.88 \\
RefCOCO testB & 22\% & 19$\times$ & 0.54 & 0.49 & 0.96 & 0.63 & 0.14 & 75\% & 50\% & 0.87 \\
\bottomrule
\end{tabularx}
\end{table}

\subsection{Effect on Token Selection}
\label{app:desink}

\begin{figure}[!b]
  \centering
  \includegraphics[width=\linewidth]{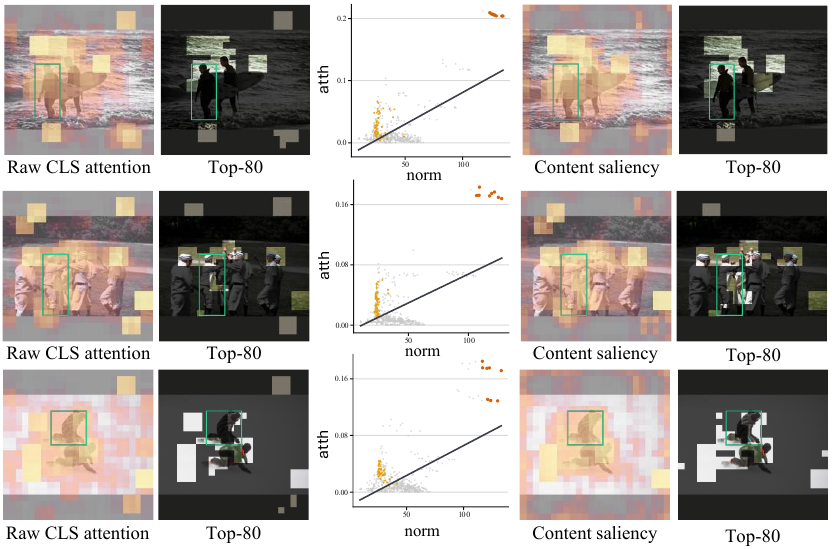}
  \caption{Three cases, one per row, five panels read left to right.
  (a)~raw CLS attention $\tilde a$; (b)~the top-$80$ cells kept
  by $\tilde a$; (c)~the $576$ cells as points, abscissa the patch norm
  $\nu$, ordinate $\tilde a$, with the least-squares baseline in dark ---
  points are \sw{routeImg}~inside the box, \sw{sinkPale}~outside it, and
  \sw{sinkTop}~the eight cells of largest norm; (d)~the residual $S$;
  (e)~the top-$80$ cells kept by $S$. In (b) and (e) kept cells retain
  their colour and discarded ones are dimmed. The \sw{gtBox}~box is the
  ground truth.}
  \label{fig:sink-case}
\end{figure}

The derivation predicts that raw CLS attention hands its top ranks to
high-norm patches, which mostly lie on the background, and that removing the
norm-explained component returns these ranks to the content.
Figure~\ref{fig:sink-case} illustrates this step from left to right on three
cases. The two
panels on the left are what one sees when raw CLS is borrowed as
a content criterion: many of the bright cells fall on sea, grass, sky and
other background, and most of the top-$80$ seats go to them. The middle
panel shows why. The $576$ points spread along a rising line --- attention
does follow norm --- and that line is the baseline to be subtracted. The
largest-norm cells are thrown far out to the right and almost all of them
lie outside the box, so their score is bought by norm. The cells inside the
box, in contrast, crowd into the low-norm side and float above the line,
and it is the part above the baseline that counts as content. Subtracting
the baseline deducts the more from a cell the larger its norm, so the
cells at the right end fall the furthest and those inside the box rise
relative to them. The two panels on the right are the result of that
subtraction: sink cells have not disappeared, but the background is
markedly weakened and the seats have returned to the target. Nor should
they disappear --- they were never pushed below the line, and a sink is
not meaningless to begin with; it aggregates information from the whole
image.

\begin{figure}[ht]
  \centering
  \includegraphics[width=\linewidth]{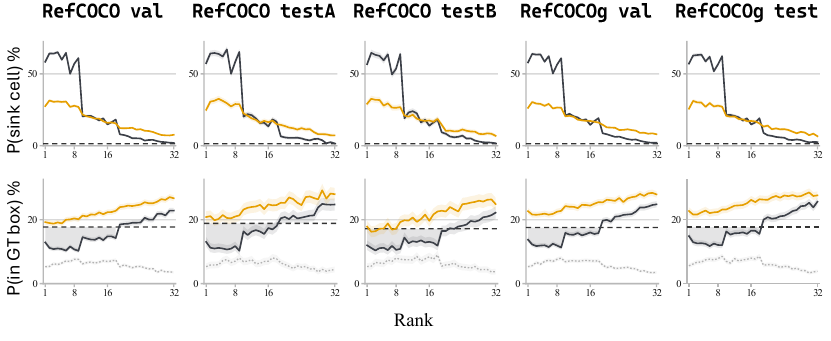}
  \caption{One column per split, $n=600$ images each. The horizontal axis
  is the rank $r$ under the score ($r=1$ is the highest-scoring cell). Top
  row: the probability that the cell ranked $r$ is one of the eight
  largest-norm cells. Bottom row: the probability that it falls inside the
  ground-truth box. Curves are \sw{sinkRaw}~raw CLS $\tilde a$
  and \sw{routeImg}~the debiased $S$, with \sw{sinkCtl}~the norm $\nu$
  itself as a control in the bottom row. Bands are $95\%$ confidence
  intervals of the mean and the dashed line is chance.}
  \label{fig:sink-prof}
  \vspace{-10pt}
\end{figure}

For token pruning a more direct reading is to ask where the high-norm
tokens rank and where the tokens inside the ground-truth box rank.
Figure~\ref{fig:sink-prof} measures both on five grounding benchmarks. The
top row plots the probability that the cell ranked $r$ is one of the eight
largest-norm cells: raw CLS hands its foremost seats to sink
cells, about six in ten of the first eight being sink cells. The bottom
row plots the probability that the cell ranked $r$ falls inside the
ground-truth box. An ideal ranking should be most accurate at $r=1$ and
decay from there, yet raw CLS starts \emph{below} chance and
climbs past it only around rank twenty, landing inside the box just
$10\%$ of the time over the first eight ranks. The norm control stays at
the bottom throughout, which confirms the premise this section rests on:
where the norm is high is background.

After debiasing both rows move in the intended direction. The peak in the
top row roughly halves and shifts right, but it does not vanish: the
median rank of a sink cell moves only from $5$--$6$ to $13$--$15$. That is
the result we want rather than a defect --- a sink still does its job of
aggregating information over the whole image, and this step only weakens
the say of the background, sink cells included. The in-box rate of the
first eight ranks rises accordingly in the bottom row, which is the direct
gain for token pruning: at the same budget, more of the selected tokens
land on the target.

\section{Algorithm for Image-Intrinsic Evidence Selection}
\label{app:cover}

Algorithm~\ref{alg:cover} shows the greedy procedure that maximises the
representational coverage in Eq.~\eqref{eq:cover}.

\begin{center}
\begin{minipage}{0.68\linewidth}
\refstepcounter{algorithm}\label{alg:cover}
\hrule height 0.8pt \vspace{3pt}
{\small\textbf{Algorithm~\thealgorithm}\; Image-intrinsic evidence by
attention-weighted coverage}
\vspace{3pt}\hrule height 0.4pt \vspace{2pt}
\small
\begin{algorithmic}[1]
\Require patch features $\{x_j\}_{j=1}^{N}$, CLS attention $\{a_j\}$,
tokens in place $\mathcal A_0$, budget $N_i$
\Ensure image-intrinsic evidence $\mathcal Q_i$ with $|\mathcal Q_i|=N_i$
\State $C_{jk}\gets\cos(x_j,x_k)$ for all $j,k$
\State $\mathrm{cov}_j\gets\max\big(0,\max_{i\in\mathcal A_0}C_{ij}\big)$
\Comment{$0$ if $\mathcal A_0=\emptyset$}
\State $\mathcal A\gets\mathcal A_0$,\;\; $\mathcal Q_i\gets\emptyset$
\For{$n=1,\dots,N_i$}
  \State $t^\star\gets\arg\max_{t\notin\mathcal A}\;\sum_j a_j\,\mathrm{relu}\big(C_{tj}-\mathrm{cov}_j\big)$
  \State $\mathcal A\gets\mathcal A\cup\{t^\star\}$,\;\; $\mathcal Q_i\gets\mathcal Q_i\cup\{t^\star\}$
  \State $\mathrm{cov}_j\gets\max\big(\mathrm{cov}_j,\,C_{t^\star j}\big)$ for all $j$
\EndFor
\State \Return $\mathcal Q_i$
\end{algorithmic}
\vspace{2pt}\hrule height 0.8pt
\end{minipage}
\end{center}

\section{Detailed Grounding Results on Qwen2.5-VL-7B}
\label{app:qwen_grounding}

Table~\ref{tab:qwen_grounding} lists the per-split grounding results behind the
REC average in Table~\ref{tab:qwen}. Bifocal is best on every split at all
three budgets.

\begin{table}[ht]
\centering
\caption{\textbf{Visual grounding on Qwen2.5-VL-7B.} All entries are
\textsc{Acc}@$0.5$.}
\label{tab:qwen_grounding}
\scriptsize
\setlength{\tabcolsep}{2.5pt}\renewcommand{\arraystretch}{0.9}
\setlength{\aboverulesep}{0pt}\setlength{\belowrulesep}{0pt}
\begin{tabularx}{\linewidth}{@{}l
  >{\hsize=1.5\hsize\centering\arraybackslash}X *{3}{>{\hsize=.75\hsize\centering\arraybackslash}X}
  >{\hsize=1.5\hsize\centering\arraybackslash}X *{2}{>{\hsize=.75\hsize\centering\arraybackslash}X}
  >{\hsize=1.5\hsize\centering\arraybackslash}X >{\hsize=.75\hsize\centering\arraybackslash}X
  >{\hsize=1.0\hsize\columncolor{black!4}\centering\arraybackslash}X@{}}
\toprule
Method & \gd{refA}{\textbf{RefCOCO} val} & \gd{refA}{test} & \gd{refA}{testA} & \gd{refA}{testB}
 & \gd{refB}{\textbf{RefCOCO+} val} & \gd{refB}{testA} & \gd{refB}{testB}
 & \gd{refC}{\textbf{RefCOCOg} val} & \gd{refC}{test} & \gd{black}{Avg.} \\
\midrule
\ub{Baseline} & \ub{89.4} & \ub{89.4} & \ub{91.6} & \ub{84.4} & \ub{83.0} & \ub{88.9} & \ub{75.8} & \ub{83.9} & \ub{84.3} & \ub{100.0\%} \\
\midrule
\rowcolor{tierbg}\multicolumn{11}{c}{\textbf{Retain 50\% Tokens}} \\
VisionZip & 77.9 & 78.4 & 81.9 & 72.9 & 75.2 & 81.1 & 68.4 & 77.5 & 77.4 & 89.6\% \\
DART & 75.7 & 76.7 & 80.8 & 71.2 & 70.0 & 76.7 & 62.8 & 72.2 & 72.8 & 85.5\% \\
N\"uwa & \underline{86.8} & \underline{87.6} & \underline{89.2} & \underline{83.0} & \underline{79.8} & \underline{85.4} & \underline{73.6} & \underline{81.9} & \underline{81.8} & \underline{97.2\%} \\
HiPrune & 77.5 & 78.0 & 81.2 & 73.2 & 75.1 & 80.6 & 68.3 & 77.5 & 77.8 & 89.4\% \\
EADP & 82.6 & 83.0 & 86.4 & 76.3 & 75.5 & 81.9 & 66.0 & 76.7 & 76.6 & 91.5\% \\
VLM-Pruner & 84.6 & 84.3 & 87.8 & 78.1 & 78.2 & 84.0 & 69.4 & 78.8 & 79.5 & 94.0\% \\
\rowcolor{oursbg} \textbf{Bifocal (ours)} & \textbf{87.8} & \textbf{88.1} & \textbf{90.8} & \textbf{83.5} & \textbf{81.4} & \textbf{86.9} & \textbf{73.7} & \textbf{83.1} & \textbf{82.6} & \textbf{98.3\%} \\
\rowcolor{tierbg}\multicolumn{11}{c}{\textbf{Retain 25\% Tokens}} \\
VisionZip & 61.5 & 61.7 & 66.6 & 54.9 & 58.5 & 65.0 & 52.0 & 59.9 & 59.0 & 69.9\% \\
DART & 26.5 & 26.6 & 29.9 & 23.3 & 24.2 & 27.9 & 20.3 & 26.2 & 25.5 & 29.9\% \\
N\"uwa & \underline{77.0} & \underline{77.6} & \underline{80.3} & \underline{73.2} & \underline{68.2} & \underline{73.9} & \underline{62.5} & \underline{72.8} & \underline{72.7} & \underline{85.4\%} \\
HiPrune & 59.4 & 60.2 & 64.4 & 53.1 & 56.4 & 63.2 & 49.2 & 59.0 & 58.4 & 67.9\% \\
EADP & 60.9 & 62.1 & 68.6 & 52.4 & 56.8 & 64.9 & 45.4 & 56.4 & 56.1 & 67.9\% \\
VLM-Pruner & 43.9 & 44.4 & 51.3 & 36.7 & 41.1 & 48.2 & 32.3 & 40.4 & 39.8 & 49.1\% \\
\rowcolor{oursbg} \textbf{Bifocal (ours)} & \textbf{81.0} & \textbf{81.8} & \textbf{84.2} & \textbf{77.8} & \textbf{72.7} & \textbf{77.2} & \textbf{66.1} & \textbf{77.1} & \textbf{77.0} & \textbf{90.2\%} \\
\rowcolor{tierbg}\multicolumn{11}{c}{\textbf{Retain 15\% Tokens}} \\
VisionZip & 38.6 & 38.9 & 44.2 & 34.1 & 36.7 & 41.6 & 30.6 & 36.6 & 36.5 & 43.8\% \\
DART & 6.0 & 6.1 & 6.3 & 6.6 & 5.2 & 5.8 & 5.4 & 6.2 & 5.5 & 6.9\% \\
N\"uwa & \underline{53.2} & \underline{53.9} & \underline{57.7} & \underline{50.1} & \underline{44.8} & \underline{48.0} & \underline{40.2} & \underline{49.3} & \underline{49.4} & \underline{58.0\%} \\
HiPrune & 35.6 & 35.9 & 39.9 & 31.6 & 32.9 & 37.3 & 28.9 & 34.6 & 34.1 & 40.3\% \\
EADP & 36.4 & 36.9 & 43.1 & 29.8 & 33.6 & 41.5 & 26.1 & 34.1 & 34.0 & 40.9\% \\
VLM-Pruner & 8.2 & 8.9 & 8.5 & 8.2 & 7.9 & 8.3 & 7.8 & 8.0 & 7.7 & 9.5\% \\
\rowcolor{oursbg} \textbf{Bifocal (ours)} & \textbf{62.1} & \textbf{63.4} & \textbf{67.2} & \textbf{58.1} & \textbf{51.5} & \textbf{58.1} & \textbf{47.8} & \textbf{57.6} & \textbf{57.3} & \textbf{67.9\%} \\
\bottomrule
\end{tabularx}
\end{table}

\section{Results on LLaVA-NeXT-7B}
\label{app:next}

The \textsc{anyres} backbone is a harder test: it splits each image into a
variable number of crops, so the visual sequence is about four times longer
than on LLaVA-1.5 and varies across images, and a fixed token budget no longer
corresponds to a fixed fraction of the sequence. As shown in
Tables~\ref{tab:next_general} and~\ref{tab:next_grounding}, the gap between
general and grounding performance widens further on this backbone. On the
general benchmarks all methods fall within a narrow band, and even at the
160-token budget the baselines differ by at most $10.9$ points in Avg., whereas
on grounding, Avg.\ ranges from $3.1\%$ to $52.1\%$. At this budget, VisionZip
reaches a general Avg.\ of $78.2\%$, close to the $79.8\%$ of Bifocal, yet
retains only $3.1\%$ on grounding, about $17\times$ lower than ours. General
VQA is therefore largely insensitive to which tokens are retained, while
grounding exposes the difference. Moreover, the four baselines that prune once
before the LLM (VisionZip, HiPrune, EADP and SID) perform similarly on
grounding at every budget, whether or not they use the query text, suggesting
that where the budget is spent in the pipeline matters more than text
awareness.

\begin{table}[!ht]
\centering
\caption{\textbf{General benchmarks on LLaVA-NeXT-7B.} Within each tier, the best and
second-best results are \textbf{bold} and \underline{underlined}. Avg.\
averages benchmark-wise performance relative to the baseline.}
\label{tab:next_general}
\scriptsize
\setlength{\tabcolsep}{1.4pt}\renewcommand{\arraystretch}{1.05}
\setlength{\aboverulesep}{0pt}\setlength{\belowrulesep}{0pt}
\begin{tabularx}{\linewidth}{@{}l *{10}{>{\centering\arraybackslash}X} >{\columncolor{black!4}\centering\arraybackslash}X@{}}
\toprule
\multirow{2}{*}{Method} & \cat{3}{catGen}{General} & \cat{2}{catRea}{Reasoning} & \cat{4}{catOcr}{OCR} & \cat{1}{catHal}{Hall.} & \cat{1}{black}{Avg.} \\
\cmidrule(lr){2-4}\cmidrule(lr){5-6}\cmidrule(lr){7-10}\cmidrule(lr){11-11}
 & GQA & MMB & VQA\textsuperscript{OK} & SQA-I & AI2D & VQA\textsuperscript{T} & OCR-B & DocVQA & ChartQA & POPE & \\
\midrule
\ub{Baseline} & \ub{64.2} & \ub{67.2} & \ub{44.2} & \ub{70.2} & \ub{65.4} & \ub{64.9} & \ub{521} & \ub{68.5} & \ub{54.9} & \ub{86.4} & \ub{100.0\%} \\
\midrule
\rowcolor{tierbg}\multicolumn{12}{c}{\textbf{Average Tokens $\approx$ 640}} \\
VisionZip & 61.2 & 65.0 & \textbf{46.8} & 68.0 & 63.5 & 61.2 & 474 & 55.3 & 42.0 & 86.2 & 93.4\% \\
SID & 62.5 & 65.1 & 40.5 & 68.3 & 63.8 & 58.1 & 415 & 50.0 & 39.5 & 86.1 & 89.5\% \\
N\"uwa & \underline{63.4} & 64.6 & 43.7 & \underline{68.8} & 64.0 & \underline{63.0} & 376 & \underline{60.9} & \underline{48.5} & 85.6 & 93.5\% \\
EADP & 63.2 & 65.5 & 39.2 & 67.0 & 64.4 & 56.8 & 405 & 47.6 & 38.3 & \textbf{87.7} & 88.4\% \\
HiPrune & 61.8 & \underline{66.2} & 41.6 & 68.6 & \textbf{65.0} & 61.6 & \underline{486} & 57.1 & 44.0 & 85.6 & \underline{93.7\%} \\
\rowcolor{oursbg} \textbf{Bifocal (ours)} & \textbf{63.9} & \textbf{67.0} & \underline{46.6} & \textbf{69.9} & \underline{64.6} & \textbf{64.2} & \textbf{507} & \textbf{65.2} & \textbf{53.2} & \underline{86.3} & \textbf{99.1\%} \\
\rowcolor{tierbg}\multicolumn{12}{c}{\textbf{Average Tokens $\approx$ 320}} \\
VisionZip & 58.9 & 63.6 & \underline{43.8} & 67.5 & 62.4 & 58.3 & 397 & 41.8 & \underline{35.9} & 82.0 & \underline{86.4\%} \\
SID & 60.6 & 63.0 & 30.8 & 67.7 & 62.1 & 53.4 & 363 & 40.8 & 34.2 & \underline{85.8} & 82.2\% \\
N\"uwa & \underline{61.9} & 63.1 & 39.8 & \textbf{68.0} & \textbf{63.8} & \underline{58.5} & 312 & \underline{49.5} & 34.7 & 83.3 & 85.7\% \\
EADP & 61.7 & 63.1 & 26.1 & 67.6 & \underline{63.2} & 49.7 & 312 & 32.6 & 27.4 & \textbf{86.1} & 77.5\% \\
HiPrune & 57.6 & \underline{64.2} & 30.2 & \underline{67.7} & 62.7 & 56.9 & \underline{402} & 40.5 & 35.8 & 79.9 & 82.8\% \\
\rowcolor{oursbg} \textbf{Bifocal (ours)} & \textbf{63.3} & \textbf{64.8} & \textbf{45.9} & 66.4 & 62.5 & \textbf{61.0} & \textbf{415} & \textbf{54.4} & \textbf{46.8} & 84.8 & \textbf{92.6\%} \\
\rowcolor{tierbg}\multicolumn{12}{c}{\textbf{Average Tokens $\approx$ 160}} \\
VisionZip & 55.5 & 59.6 & \textbf{41.6} & \underline{68.1} & \underline{62.0} & \underline{52.6} & \textbf{308} & 27.5 & \underline{29.6} & 74.7 & \underline{78.2\%} \\
SID & 57.7 & 58.9 & 18.1 & 67.4 & 61.4 & 46.6 & 283 & 28.0 & 24.7 & \underline{82.7} & 71.6\% \\
N\"uwa & 58.5 & \underline{60.1} & 33.3 & 66.5 & \textbf{62.2} & 51.8 & 219 & \underline{31.4} & 28.1 & 77.6 & 75.4\% \\
EADP & \underline{59.2} & 59.8 & 16.3 & 67.1 & 61.1 & 39.2 & 253 & 22.6 & 22.3 & 82.3 & 68.5\% \\
HiPrune & 51.6 & 59.9 & 17.5 & \textbf{68.7} & 60.7 & 42.7 & 259 & 22.0 & 26.8 & 65.9 & 67.3\% \\
\rowcolor{oursbg} \textbf{Bifocal (ours)} & \textbf{60.4} & \textbf{60.8} & \underline{38.6} & 59.7 & 58.8 & \textbf{54.4} & \underline{298} & \textbf{36.3} & \textbf{33.1} & \textbf{83.8} & \textbf{79.8\%} \\
\bottomrule
\end{tabularx}
\end{table}

\begin{table}[!ht]
\centering
\caption{\textbf{Visual grounding on LLaVA-NeXT-7B (\textmd{\textsc{anyres}}).} All
entries are \textsc{Acc}@$0.5$.}
\label{tab:next_grounding}
\scriptsize
\setlength{\tabcolsep}{2.5pt}\renewcommand{\arraystretch}{0.9}
\setlength{\aboverulesep}{0pt}\setlength{\belowrulesep}{0pt}
\begin{tabularx}{\linewidth}{@{}l
  >{\hsize=1.5\hsize\centering\arraybackslash}X *{2}{>{\hsize=.7\hsize\centering\arraybackslash}X}
  >{\hsize=1.5\hsize\centering\arraybackslash}X *{2}{>{\hsize=.7\hsize\centering\arraybackslash}X}
  >{\hsize=1.5\hsize\centering\arraybackslash}X >{\hsize=.7\hsize\centering\arraybackslash}X
  >{\hsize=1.0\hsize\columncolor{black!4}\centering\arraybackslash}X@{}}
\toprule
Method & \gd{refA}{\textbf{RefCOCO} val} & \gd{refA}{testA} & \gd{refA}{testB}
 & \gd{refB}{\textbf{RefCOCO+} val} & \gd{refB}{testA} & \gd{refB}{testB}
 & \gd{refC}{\textbf{RefCOCOg} val} & \gd{refC}{test} & \gd{black}{Avg.} \\
\midrule
\ub{Baseline} & \ub{85.4} & \ub{89.6} & \ub{78.0} & \ub{77.0} & \ub{86.4} & \ub{66.4} & \ub{81.9} & \ub{80.4} & \ub{100.0\%} \\
\midrule
\rowcolor{tierbg}\multicolumn{10}{c}{\textbf{Average Tokens $\approx$ 640}} \\
VisionZip & 8.5 & 8.4 & 8.2 & 7.1 & 7.9 & 7.0 & 9.0 & 9.2 & 10.1\% \\
SID & 35.0 & 38.9 & 30.6 & 30.8 & 34.8 & 26.0 & 33.1 & 33.5 & 40.7\% \\
N\"uwa & \underline{63.9} & \underline{67.7} & \underline{57.4} & \underline{56.3} & \underline{63.2} & \underline{46.8} & \underline{57.7} & \underline{56.9} & \underline{72.8\%} \\
EADP & 35.9 & 39.0 & 30.9 & 31.4 & 35.1 & 25.1 & 33.8 & 34.2 & 41.1\% \\
HiPrune & 32.0 & 35.3 & 29.5 & 27.2 & 31.5 & 24.2 & 29.0 & 29.1 & 36.9\% \\
\rowcolor{oursbg} \textbf{Bifocal (ours)} & \textbf{78.1} & \textbf{83.6} & \textbf{69.6} & \textbf{70.2} & \textbf{79.2} & \textbf{58.2} & \textbf{73.9} & \textbf{72.5} & \textbf{90.7\%} \\
\rowcolor{tierbg}\multicolumn{10}{c}{\textbf{Average Tokens $\approx$ 320}} \\
VisionZip & 4.5 & 4.2 & 4.8 & 3.7 & 3.5 & 4.2 & 4.7 & 4.8 & 5.3\% \\
SID & 21.9 & 23.8 & 18.7 & 20.1 & 21.9 & 16.1 & 20.3 & 20.1 & 25.2\% \\
N\"uwa & \underline{33.7} & \underline{35.4} & \underline{32.8} & \underline{26.1} & \underline{29.6} & \underline{22.5} & \underline{28.6} & \underline{28.4} & \underline{36.8\%} \\
EADP & 22.6 & 24.6 & 19.4 & 21.1 & 22.6 & 17.9 & 22.3 & 21.9 & 26.7\% \\
HiPrune & 13.2 & 11.2 & 13.0 & 11.7 & 10.6 & 11.5 & 11.6 & 12.7 & 14.8\% \\
\rowcolor{oursbg} \textbf{Bifocal (ours)} & \textbf{64.7} & \textbf{70.7} & \textbf{56.7} & \textbf{56.1} & \textbf{65.3} & \textbf{45.1} & \textbf{59.9} & \textbf{58.7} & \textbf{74.0\%} \\
\rowcolor{tierbg}\multicolumn{10}{c}{\textbf{Average Tokens $\approx$ 160}} \\
VisionZip & 2.6 & 2.1 & 3.3 & 2.0 & 1.6 & 2.7 & 2.7 & 2.8 & 3.1\% \\
SID & 12.4 & 12.3 & 12.2 & 11.1 & 11.5 & 10.6 & 10.6 & 11.5 & 14.3\% \\
N\"uwa & \underline{17.9} & 17.3 & \underline{18.4} & 12.5 & 13.1 & 12.7 & \underline{15.2} & \underline{15.5} & \underline{19.0\%} \\
EADP & 16.0 & \underline{17.4} & 14.5 & \underline{15.1} & \underline{15.9} & \underline{13.0} & 14.6 & 14.7 & 18.8\% \\
HiPrune & 5.0 & 4.2 & 5.6 & 5.0 & 3.8 & 5.1 & 4.8 & 4.7 & 5.9\% \\
\rowcolor{oursbg} \textbf{Bifocal (ours)} & \textbf{45.6} & \textbf{50.4} & \textbf{40.2} & \textbf{38.3} & \textbf{45.5} & \textbf{31.2} & \textbf{42.9} & \textbf{42.1} & \textbf{52.1\%} \\
\bottomrule
\end{tabularx}
\end{table}
\section{Efficiency Analysis}
\label{app:efficiency}

Token pruning primarily reduces prefill computation, but FLOP reductions
do not necessarily translate into proportional speedups, as practical
efficiency also depends on execution patterns and kernel
implementations~\citep{hu2026beyond}.
We therefore report FLOPs, per-sample prefill time, time to first token
(TTFT), and peak memory on TextVQA at the $128$-token budget in
Table~\ref{tab:efficiency}. While leading clearly on
accuracy, Bifocal also achieves a $1.27\times$ speedup in TTFT, faster than
EADP, DART and VLM-Pruner and close to the fastest methods, with no extra
memory cost.

\begin{table}[!ht]
\centering
\caption{\textbf{Efficiency on LLaVA-1.5-7B.}}
\label{tab:efficiency}
\footnotesize
\setlength{\tabcolsep}{4pt}\renewcommand{\arraystretch}{1.15}
\setlength{\aboverulesep}{0pt}\setlength{\belowrulesep}{0pt}
\begin{tabularx}{\linewidth}{@{}l *{2}{>{\centering\arraybackslash}X} c *{3}{>{\centering\arraybackslash}X}@{}}
\toprule
Method & Avg.\ Tokens & FLOPs (T) & Peak Mem.\ (GiB) & Prefill (ms) & TTFT (ms) & Speedup \\
\midrule
LLaVA-1.5-7B & 576 & 8.450 & 13.5 & 50.36 & 89.32 & 1.00$\times$ \\
\midrule
VLM-Pruner & 128 & 2.459 & 13.8 & 45.54 & 85.57 & 1.04$\times$ \\
EADP & 128 & 2.454 & 13.5 & 25.41 & 81.67 & 1.09$\times$ \\
DART & 128 & 2.483 & 13.4 & 35.86 & 75.00 & 1.19$\times$ \\
VisionZip & 128 & 2.454 & 13.5 & 24.37 & 65.42 & 1.37$\times$ \\
HiPrune & 128 & 2.454 & 13.5 & 24.40 & 64.75 & 1.38$\times$ \\
\rowcolor{oursbg} \textbf{Bifocal (ours)} & 128 & 2.458 & 13.4 & 28.61 & 70.61 & 1.27$\times$ \\
\bottomrule
\end{tabularx}
\end{table}

\section{Details of Benchmarks}
\textbf{VQAv2}~\citep{goyal2017making}. VQAv2 evaluates open-ended visual question answering on balanced image--question pairs, with complementary images designed to reduce language priors and encourage stronger visual grounding.

\textbf{GQA}~\citep{hudson2019gqa}.
GQA evaluates compositional visual reasoning over natural images using questions derived from scene graphs, emphasizing objects, attributes, spatial relations, and structured reasoning.

\textbf{MMBench}~\citep{liu2024mmbench}.
MMBench is a comprehensive multimodal benchmark that evaluates a broad range of perception and reasoning abilities through carefully designed multiple-choice questions.

\textbf{OK-VQA}~\citep{marino2019ok}.
OK-VQA evaluates knowledge-based visual question answering, where answering often requires external or commonsense knowledge beyond information directly observable in the image.

\textbf{ScienceQA}~\citep{lu2022learn}.
ScienceQA evaluates multimodal scientific reasoning over questions from natural science, social science, and language domains, requiring models to combine visual understanding with contextual and domain knowledge.

\textbf{AI2D}~\citep{kembhavi2016diagram}.
AI2D evaluates diagram understanding and reasoning over scientific illustrations, requiring models to interpret visual components, textual labels, symbols, and their structural relationships.

\textbf{TextVQA}~\citep{singh2019towards}.
TextVQA evaluates visual question answering over text-rich natural images, requiring models to recognize scene text and reason jointly over textual and surrounding visual information.

\textbf{OCRBench}~\citep{liu2024ocrbench}.
OCRBench evaluates OCR-related capabilities of multimodal large language models across text recognition, scene-text understanding, document understanding, and information extraction.

\textbf{DocVQA}~\citep{mathew2021docvqa}.
DocVQA evaluates question answering over document images, emphasizing text recognition, layout understanding, information localization, and reasoning over structured document content.

\textbf{ChartQA}~\citep{masry2022chartqa}.
ChartQA evaluates question answering over chart images, requiring models to understand chart elements and perform visual, logical, and arithmetic reasoning over the underlying data.

\textbf{POPE}~\citep{li2023evaluating}.
POPE is a diagnostic benchmark for object hallucination, using binary questions to test whether multimodal models incorrectly report objects that are absent from the image.

\textbf{RefCOCO}~\citep{kazemzadeh2014referitgame,yu2016modeling}.
RefCOCO evaluates referring expression comprehension by requiring models to localize objects described by natural-language expressions in complex visual scenes.

\textbf{RefCOCO+}~\citep{kazemzadeh2014referitgame,yu2016modeling}.
RefCOCO+ evaluates referring expression comprehension while restricting explicit location words, placing greater emphasis on object appearance, attributes, and relational cues.

\textbf{RefCOCOg}~\citep{kazemzadeh2014referitgame,yu2016modeling}.
RefCOCOg contains longer and more descriptive referring expressions, requiring models to integrate fine-grained object attributes and contextual relationships for accurate visual grounding.

\clearpage
\section{Visualisations}
\label{app:visual}


\subsection{Fine-Grained Tasks on Qwen2.5-VL-7B}

\begin{figure}[!ht]
  \centering
  \includegraphics[width=\linewidth]{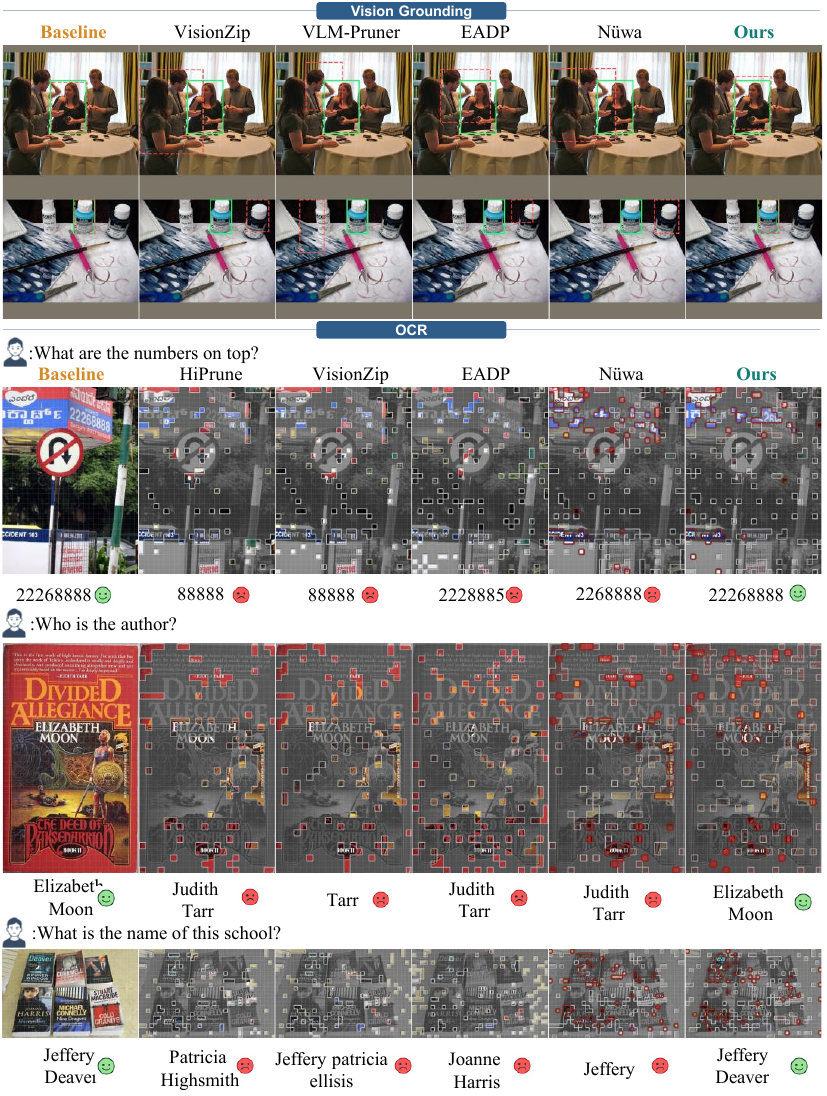}
  \caption{\textbf{Grounding and OCR cases on Qwen2.5-VL-7B.} In the grounding
  cases, \mkgt~is the ground truth and \mkpred~the prediction.}
  \label{fig:visual-qwen}
\end{figure}




\vspace{-6pt}
\subsection{Fine-Grained Tasks on LLaVA-1.5-7B}
{\centering\captionsetup{type=figure}
  \includegraphics[width=\linewidth,height=\textheight,keepaspectratio]%
    {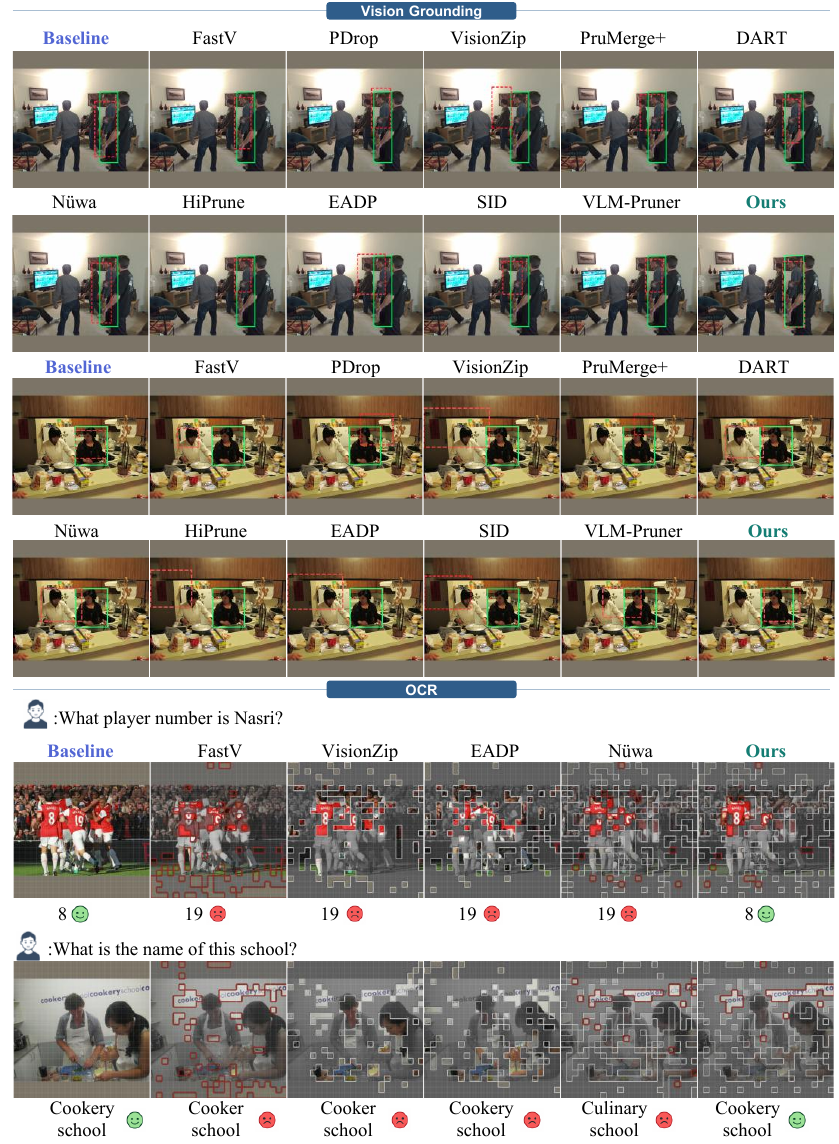}
  \caption{\textbf{Grounding and OCR cases on LLaVA-1.5-7B.} Outputs of
  the unpruned model and of pruning methods under the same budget.}
  \label{fig:visual-llava}\par}

\clearpage

\subsection{Image-Intrinsic and Query-Focused Evidence}

\begin{figure}[htbp]
  \centering
  \includegraphics[width=\linewidth]{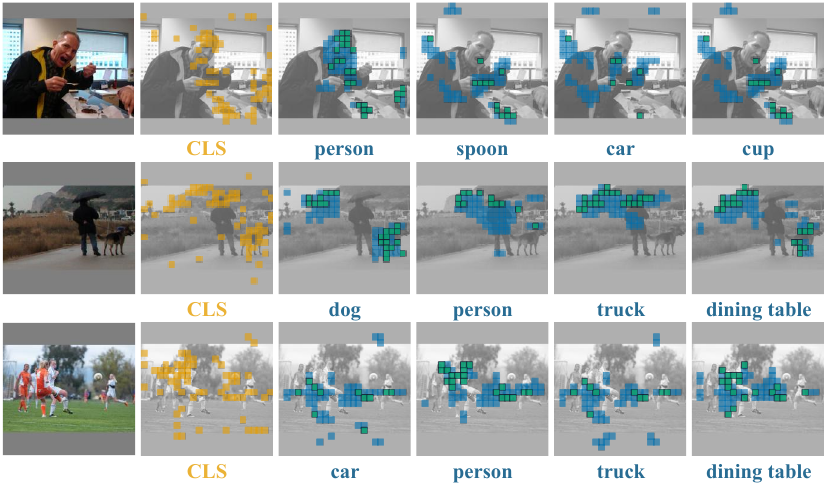}
  \caption{\textbf{The two kinds of fine evidence under different queries.}
  Each row is one POPE image. The first panel after the input shows
  \sw{routeImg}~the image-intrinsic evidence, and the remaining panels show
  \sw{routeTxt}~the query-focused evidence for each query.}
  \label{fig:visual-routes}
\end{figure}

Figure~\ref{fig:visual-routes} shows the fine evidence Bifocal selects for one
image under several POPE queries. The image-intrinsic evidence, selected from
the CLS attention, spreads over the whole image and barely changes across
queries. The query-focused evidence instead changes with the query and gathers
on the object it names. The two kinds of evidence thus play the roles they are
designed for: one covers the image, the other follows the question.

\end{document}